\documentclass{article}

\usepackage{microtype}
\usepackage{graphicx}
\usepackage{subcaption}
\usepackage{booktabs} 

\usepackage[hyphens]{url}
\usepackage{hyperref}
\usepackage[dvipsnames]{xcolor}
\usepackage{comment}
\usepackage{stfloats}

\newcommand{\signcolor}[1]{%
  \ifdim #1 pt < 0pt
    {\bfseries\color{Green} #1}%
  \else\ifdim #1 pt = 0pt
    {\color{Black} #1}%
  \else
    {\bfseries\color{Red} #1}%
  \fi\fi
}

\usepackage[preprint]{icml2026}

\usepackage{amsmath}
\usepackage{amssymb}
\usepackage{mathtools}
\usepackage{amsthm}
\usepackage{multirow}
\usepackage{multicol}

\usepackage[capitalize,noabbrev]{cleveref}

\theoremstyle{plain}

\theoremstyle{definition}

\theoremstyle{remark}

\usepackage[textsize=tiny]{todonotes}

\usepackage{xspace} 
\newcommand{\modulenameFull}{Modular Angular-Radial Attention}
\newcommand{\modulenameShort}{MARA\xspace}
\newcommand{\maceModulename}{MACE-MARA\xspace}

\icmltitlerunning{\modulenameShort: Continuous SE(3)-Equivariant Attention for Molecular Force Fields}

\usepackage{physics}

\begin{document}

\twocolumn[
  \icmltitle{\modulenameShort: Continuous SE(3)-Equivariant Attention for Molecular Force Fields}



  \icmlsetsymbol{equal}{*}

  \begin{icmlauthorlist}
    \icmlauthor{Francesco Leonardi}{yyy}
    \icmlauthor{Boris Bonev}{comp}
    \icmlauthor{Kaspar Riesen}{yyy}
  \end{icmlauthorlist}

  \icmlaffiliation{yyy}{Institute of Computer Science, University of Bern, 3012 Bern, Switzerland}
  \icmlaffiliation{comp}{NVIDIA Corporation, 95051 Santa Clara, CA, USA}
  
  \icmlcorrespondingauthor{Francesco Leonardi}{francesco.leonardi@unibe.ch}

  \icmlkeywords{Graph Neural Network, Machine Learning Force Field, Spherical Attention}

  \vskip 0.3in
]

\printAffiliationsAndNotice{}  


\begin{abstract}
    Machine learning force fields (MLFFs) have become essential for accurate and efficient atomistic modeling. Despite their high accuracy, most existing approaches rely on fixed angular expansions, limiting flexibility in weighting local geometric interactions. We introduce Modular Angular-Radial Attention (MARA), a module that extends spherical attention -- originally developed for SO(3) tasks -- to the molecular domain and SE(3), providing an efficient approximation of equivariant interactions. MARA operates directly on the angular and radial coordinates of neighboring atoms, enabling flexible, geometrically informed, and modular weighting of local environments. Unlike existing attention mechanisms in SE(3)-equivariant architectures, MARA can be integrated in a plug-and-play manner into models such as MACE without architectural modifications. Across molecular benchmarks, MARA improves energy and force predictions, reduces high-error events, and enhances robustness. These results demonstrate that continuous spherical attention is an effective and generalizable geometric operator that increases the expressiveness, stability, and reliability of atomistic models.
\end{abstract}

\section{Introduction}

\begin{figure*}[ht]
	\vskip 0.2in
	\begin{center}
        \centerline{\includegraphics[width=\linewidth]{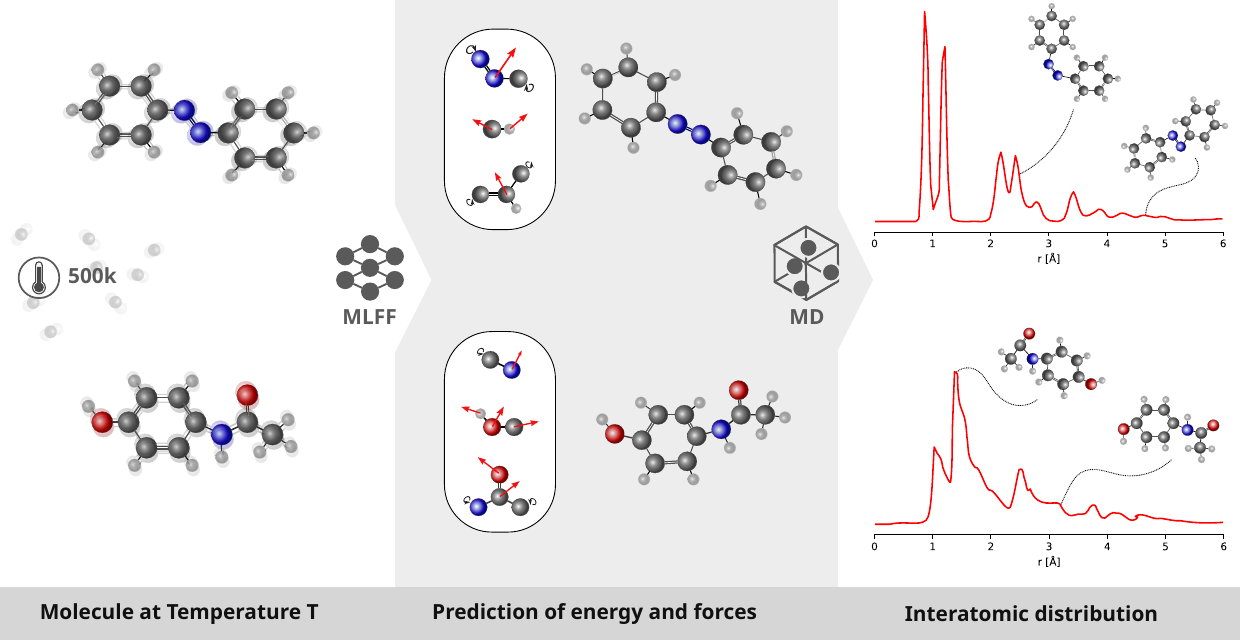}}
		\caption{
			Illustration of a Machine Learning Force Field (MLFF) modeled with Equivariant Graph Neural Networks (EGNNs). Given a molecular configuration at a specific temperature, the model predicts the forces acting on each atom, which are then used to estimate the molecule’s subsequent configuration. Molecular configurations that occur more frequently are captured more accurately by the model, while less frequent configurations, which can be visualized through the interatomic distribution, are inherently more uncertain.
		}
		\label{img:intro}
	\end{center}    
	\vskip -0.2in
\end{figure*}

Machine Learning Force Fields (MLFFs) have become a central tool for atomistic simulations in chemistry and materials science~\cite{doi:10.1021/acs.jpcc.6b10908, doi:10.1021/acs.chemrev.0c01111}, offering an effective compromise between the accuracy of \emph{ab initio} electronic structure methods, such as Density Functional Theory (DFT), and the computational efficiency required for large-scale simulations~\cite{doi:10.1021/acs.jpca.0c04526, chmiela2023accurate}. Although early MLFF approaches included neural networks~\cite{behler2007generalized} or fixed parametric forms~\cite{THOMPSON2015316, doi:10.1137/15M1054183}, recent advances leverage SE(3)-equivariant graph neural networks to encode physical symmetries and capture complex geometric interactions\cite{DBLP:journals/corr/abs-2104-13478}. Models such as NequIP~\cite{DBLP:journals/corr/abs-2101-03164} and MACE~\cite{Batatia2022mace} trained on high-quality quantum datasets, achieve near-DFT accuracy and the scalability required for larger systems and longer simulations~\cite{behler2016perspective, chmiela2018towards}.

Despite their expressivity, most SE(3)-equivariant model angular dependencies using fixed basis expansions or discrete neighbor sets, limiting flexibility in weighting continuous directional contributions~\cite{DBLP:journals/corr/abs-2101-03164}. Inspired by the success of attention mechanisms~\cite{DBLP:conf/nips/VaswaniSPUJGKP17} in other domains, recent work has proposed SE(3)-equivariant Transformers~\cite{DBLP:conf/iclr/LiaoS23, DBLP:conf/iclr/LiaoWDS24}, but these typically operate over discrete atomic neighborhoods rather than the continuous spatial domain.

A complementary approach leverages attention directly on the sphere, providing efficient approximations to SO(3)-equivariant operators and enabling flexible modeling of angular dependencies without relying on high-order tensor products~\cite{attention_on_sphere}. Such spherical attention mechanisms have achieved strong results in geometric tasks, including spherical image segmentation and physical simulations~\cite{DBLP:journals/corr/abs-2202-04942, DBLP:conf/cvpr/BennyW25, attention_on_sphere}. Despite their advantages, they have not been explored in the context of atomistic modeling. In particular, it is unclear how these approaches can be extended to approximate SE(3) equivariance -- requiring the joint treatment of angular and radial degrees of freedom -- or how they can be integrated into existing MLFF architectures without compromising computational efficiency.

\paragraph{Our contributions}
In this work, we introduce a continuous spherical attention mechanism that operates directly on local atomic environments and provides an efficient approximation to SE(3)-equivariance. Our approach reformulates attention as an operator over directions and distances, using a structured spherical discretization that naturally separates angular and radial dependencies. This formulation enables flexible, geometry-aware weighting of local atomic environments. Moreover, we demonstrate a plug-and-play integration of this module into existing SE(3)-equivariant MLFFs, instantiated in MACE to validate its integration, while remaining fully compatible with SE(3)-equivariant message passing. Finally, we evaluate the effectiveness of our approach on representative molecular benchmarks, showing consistent improvements over current baselines with a controlled computational overhead.

\begin{figure*}[ht]
	\begin{center}
		\centerline{\includegraphics[width=0.96\linewidth]{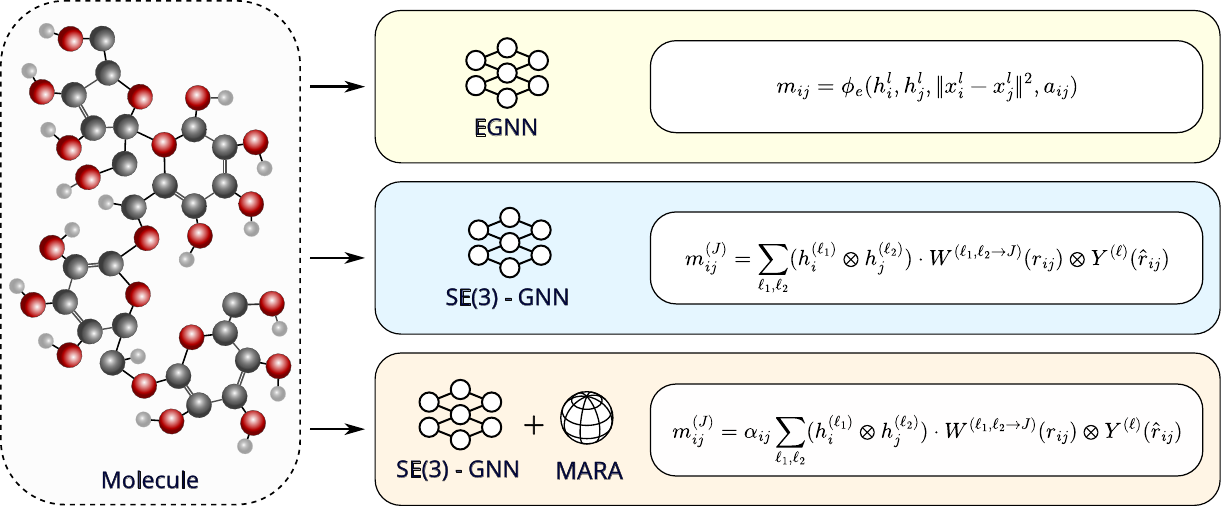}}
		\caption{
			The molecule is represented as a 3D molecular graph, in which the nodes are characterized by scalar features and spatial coordinates. Different equivariant graph neural networks process this information using different message passing mechanisms: Equivariant Graph Neural Networks (EGNN), SE(3)-equivariant networks (SE(3)-GNN), and SE(3)-GNN enriched with MARA.
		}
		\label{img:methods}
	\end{center}
	\vskip -0.2in
\end{figure*}

\section{Related Work}

\paragraph{Equivariant graph neural networks for MLFFs.}
Early MLFFs relied on neural networks~\cite{behler2016perspective} or fixed parametric forms~\cite{bartok2010gaussian, THOMPSON2015316, doi:10.1137/15M1054183}, achieving moderate accuracy and scalability. Graph-based models such as SchNet~\cite{DBLP:conf/nips/SchuttKFCTM17} and DimeNet~\cite{DBLP:conf/iclr/KlicperaGG20} are invariant to translations, rotations, and atom permutations. More recent approaches explicitly enforce SE(3) equivariance~\cite{DBLP:journals/corr/abs-2104-13478, duval2023hitchhiker}. These include Tensor Field Network (TFN)-based architectures~\cite{DBLP:journals/corr/abs-1802-08219} as well as other SE(3)-equivariant graph neural networks, such as the E(n)-equivariant GNN~\cite{DBLP:conf/icml/SatorrasHW21}. Prominent SE(3)-equivariant message-passing models include NequIP~\cite{DBLP:journals/corr/abs-2101-03164}, Allegro~\cite{DBLP:journals/corr/abs-2204-05249}, BOTNet~\cite{Batatia2022Design}, MACE~\cite{Batatia2022mace}, UNiTE~\cite{doi:10.1073/pnas.2205221119}, QuinNet~\cite{DBLP:conf/nips/WangLZWS23}, and VisNet~\cite{wang2024enhancing}, which leverage equivariant message passing and higher-order geometric representations to achieve strong performance across molecular and materials benchmarks. While highly expressive, many of these architectures rely on fixed angular basis expansions or predefined correlation orders, which can increase model complexity when modeling flexible angular interactions.

\paragraph{Transformer-based approaches in MLFFs.}
Motivated by the success of attention-based models in other domains~\cite{DBLP:conf/nips/VaswaniSPUJGKP17}, several works have introduced SE(3)-equivariant Transformer architectures, including the SE(3)-Transformer~\cite{DBLP:conf/nips/FuchsW0W20}, Equiformer~\cite{DBLP:conf/iclr/LiaoS23}, EquiformerV2~\cite{DBLP:conf/iclr/LiaoWDS24}, and SO3krates~\cite{frank2024euclidean}. These models apply attention mechanisms over local atomic neighborhoods while preserving equivariance through tensor representations and spherical harmonics. Attention is defined over discrete sets of neighboring atoms, with angular dependencies handled explicitly via spherical-harmonic–based equivariant features, rather than through attention operators acting continuously over the angular domain. In contrast, Equivariant Spherical Transformer~\cite{DBLP:journals/corr/abs-2505-23086} applies a Transformer-like architecture in the spherical Fourier domain aiming approximate equivariance via uniform spherical sampling.

\paragraph{Spherical attention mechanisms.}
Recent work has explored attention mechanisms defined directly on the sphere~\cite{attention_on_sphere}. Such spherical Transformers have demonstrated strong performance in geometric tasks, including simulations of shallow water equations and spherical image segmentation~\cite{attention_on_sphere}. By operating on spherical signals, these methods provide an efficient way to model angular dependencies and to construct operators with rotational equivariance, without explicitly relying on high-order irreducible tensor representations. To our knowledge, attention on the sphere has not yet been applied to atomistic modeling, providing an opportunity to extend these ideas to SE(3)-equivariant MLFFs, which we explore in this work.

\section{Method}

\paragraph{Graph representation.}
A molecule is represented as a graph $G = (V, E)$, where $V = \{1, \dots, N\}$ denotes the set of nodes corresponding to atoms, and $E \subseteq V \times V$ denotes the set of edges encoding pairwise interactions, such as chemical bonds or neighborhood relations.
Each node $i \in V$ is associated with a feature vector $\mathbf{h}_i \in \mathbb{R}^{d_h}$, while each edge $(i,j) \in E$ may be associated with an edge feature vector $\mathbf{e}_{ij} \in \mathbb{R}^{d_e}$. Edge features may encode bond types, interatomic distances, or other geometric descriptors.

\paragraph{3D molecular graphs.}
In addition to node and edge features, each node $i \in V$ is associated with a three-dimensional coordinate $\mathbf{x}_i \in \mathbb{R}^3$, representing the atomic position in Euclidean space.
Edges are typically defined either by spatial proximity, e.g., connecting nodes within a cutoff radius $r > 0$.

\paragraph{Geometric symmetries.}
Molecular systems are equivariant to rigid motions in three-dimensional space.
We consider the action of the special Euclidean group $\mathrm{SE}(3)$ on the atomic coordinates, such that for any $(R, \mathbf{t}) \in \mathrm{SE}(3)$,
\begin{equation}
\mathbf{x}_i \mapsto R \mathbf{x}_i + \mathbf{t}.
\end{equation}
Node and edge features are assumed to transform as scalars under $\mathrm{SE}(3)$, and therefore remain invariant. Models operating on 3D molecular graphs should preserve $\mathrm{SE}(3)$-equivariance for vectorial quantities (such as forces), with invariance enforced at the output level for scalar target properties, such as energies or molecular descriptors.

\paragraph{Attention on the sphere.}
We base our method on the continuous-domain attention mechanism, which can be understood as a generalized kernel regression over query, key, and value signals. For inputs $q,k : S^2 \to \mathbb{R}^d$ and $v : S^2 \to \mathbb{R}^e$, the spherical attention operator is defined as
\begin{multline}
\mathrm{Attn}_{S^2}[q,k,v](x) \\ 
=
\frac{\int_{S^2} \exp\big(q(x)^\top k(x') / \sqrt{d}\big) v(x') \, d\mu(x')}
{\int_{S^2} \exp\big(q(x)^\top k(x') / \sqrt{d}\big) \, d\mu(x')},
\end{multline}

where $\mu$ denotes the invariant Haar measure on the sphere. This construction ensures that the attention is equivariant under 3D rotations $R \in \mathrm{SO}(3)$:
\begin{equation}
    \mathrm{Attn}_{S^2}[q,k,v](R^{-1}x) = \mathrm{Attn}_{S^2}[q',k',v'](x). 
\end{equation}

To implement the operator numerically, \citet{attention_on_sphere} discretize the spherical domain using a set of grid points $\{g_k \in S^2\}_{k=1}^{N_\mathrm{grid}}$ and associated quadrature weights $\omega_i$, such that
\begin{equation}
    \int_{S^2} u(x) \, d\mu(x) \approx \sum_{k=1}^{N_\mathrm{grid}} u(g_k) \, \omega_k.
\end{equation}
This yields a discrete attention mechanism that preserves approximate SO(3) equivariance, enabling consistent evaluation under 3D rotations.

\paragraph{Attention SE(3)-equivariant approximation for molecular graphs.}
While the spherical attention operator is approximately equivariant under rotations SO(3), molecular coordinates also undergo translations in $\mathbb{R}^3$. To capture both directional and distance-dependent interactions, we associate with each atom $i \in V$ a discrete spherical grid $\{g_k \in S^2\}_{k=1}^{N_\mathrm{grid}}$ and define a scalar field $f_k : S^2 \to \mathbb{R}$, discretized on the spherical grid, encoding the Euclidean distances to its neighboring atoms $j \in \mathcal{N}(i)$. Concretely, for each neighbor $j$, we set
\begin{equation}
f_{jk}(g_k) = \big\lVert \mathbf{x}_j - (\mathbf{x}_i + r_{jk} \, g_k) \big\rVert
\end{equation}
where $r_{jk} = \lVert \mathbf{x}_j - \mathbf{x}_i \rVert$ and $g_k \in S^2 \text{ (unit vectors)}$. This construction guarantees that the scalar field attains its minimum along the direction pointing to each neighboring atom. Unlike models such as SchNet~\cite{DBLP:conf/nips/SchuttKFCTM17}, which rely solely on interatomic distances, our spherical scalar field encodes both directional and distance information, enabling the network to distinguish geometrically distinct configurations.

\paragraph{Spherical positional embeddings.}
Following the idea of the original work~\cite{attention_on_sphere}, we associate each point of the spherical grid $g_k$ with a learnable positional embedding $p_k$.

Atom-centered queries, keys, and values are obtained via separate linear projections of the node features, the spherical scalar field, and the positional embeddings, which are then summed:
\begin{align*}
    \mathbf{q}_{i,j,k} &= W_h \mathbf{h}_i + W_f f_{jk}(g_k) + p_k, \\
    \mathbf{k}_{i,j,k} &= W'_h \mathbf{h}_j + W'_f f_{jk}(g_k) + p_k, \\
    \mathbf{v}_{i,j,k} &= W''_h \mathbf{h}_j + W''_f f_{jk}(g_k) + p_k.
\end{align*}
where $W_h, W'_h, W''_h \in \mathbb{R}^{d \times d_h}$ and
$W_f, W'_f, W''_f \in \mathbb{R}^{d \times d_f}$ are learnable linear mappings. For simplicity, we choose the value dimension to coincide with the query/key dimension, i.e. $e = d$. This approach mirrors standard Transformer positional embeddings, allowing the attention to be aware of spherical directions while maintaining approximate rotation equivariance in the spherical domain and translation invariance through relative distances.

\paragraph{Integration into SE(3)-equivariant message passing.}
The spherical attention module is integrated as a scalar interaction block within existing SE(3)-equivariant architectures.
In particular, we apply it in parallel to the standard equivariant message passing layers of MACE~\cite{Batatia2022mace}. The choice of MACE is motivated by its widespread adoption in the current state of the art of machine learning force fields, where its architecture serves as the backbone for several foundation models~\cite{osti_3004130}. In standard SE(3)-equivariant models such as MACE, atom $i$ updates its features by aggregating messages from neighbors $j \in \mathcal{N}(i)$:
\begin{equation}
\tilde{m}_{i,j}^{(J)} = \sum_{\ell_1,\ell_2} (h_i^{(\ell_1)} \otimes h_j^{(\ell_2)}) \cdot W^{(\ell_1,\ell_2 \to J)}(r_{ij}) \otimes Y^{(\ell)}(\hat{r}_{ij}),
\end{equation}
where $W^{(\ell_1,\ell_2 \to J)}$ combines radial functions, spherical harmonics, and Clebsch--Gordan coefficients.

At each interaction layer, the backbone computes equivariant messages
$\tilde{m}_{ij,J}^{(t)}$ using tensor products of irreducible representations. The gating coefficient $\alpha_{ij}$ is obtained from the output of the spherical attention operator by applying a readout over channels and spherical grid points:

{
\vskip -0.1in
\begin{equation}
    \begin{aligned}
    \alpha_{ij}
    &= \\
    & \sigma\!\Big(
    W_{\mathrm{gate}}
    \;
    \mathrm{Pool}_{k,c} \big[
    \mathrm{Attn}_{S^2}
    [
    \mathbf{q}_{i,j,k},
    \mathbf{k}_{i,j,k},
    \mathbf{v}_{i,j,k}
    ]
    (g_k)
    \big]
    \Big)
    \end{aligned}
\end{equation}
}

where $\mathrm{Pool}_{k,c}$ indicates a rotation-invariant pooling over spherical directions $k$ and attention feature channels $c$. The final message update is then given by
\begin{equation}
m_{i,J}^{(t)} =
\sum_{j \in \mathcal{N}(i)}
\alpha_{ij} \, \tilde{m}_{ij,J}^{(t)}.
\end{equation}
Since $\alpha_{ij}$ is a scalar, equivariance is preserved only if $\alpha_{ij}$ is invariant with respect to rotations and translations. Due to the discrete approximation of the continuous attention mechanism, the resulting operation is only approximately SE(3)-equivariant. 

This formulation enables direction-dependent weighting of local atomic environments while remaining fully compatible with existing equivariant message passing schemes. When $\alpha_{ij}=1$, the formulation reduces exactly to the original MACE update, highlighting the plug-and-play nature of the proposed module.

Thus, by combining the scalar spherical attention module with the MACE backbone, we obtain an enhanced approximately SE(3)-equivariant message passing architecture, which we refer to as \maceModulename


\begin{table*}[t]
	\caption{Performance on the rMD17 dataset. Results are reported in terms of mean absolute error (MAE). Energies (E) and forces (F) are measured in kcal/mol and kcal/mol/\AA, respectively. The best results are highlighted in bold, while the second-best results are underlined. Results of the models are taken from \citet{DBLP:conf/nips/WangLZWS23}.}
	\label{table:ablation_SOTA}
	\begin{center}
		\begin{small}
			\begin{sc}				
				\resizebox{0.96\linewidth}{!}{
				\begin{tabular}{rc ccccccccr}
					\toprule
					\multicolumn{8}{c}{ } & \multicolumn{3}{c}{MACE} \\
					\cmidrule(lr){9-11}
					Molecule & & UNiTE & NequIP & Allegro & BOTNet & VisNet & QuinNet & Baseline & \modulenameShort & $\Delta$ (\%)\\
					\midrule
					
					\multirow{2}{*}{Aspirin} 
					& E & 0.055 & 0.0530 & 0.0507 & 0.0530 & \underline{0.0445} & 0.0486 & 0.0507 & \textbf{0.0419} & \signcolor{-17.36}\\					
					& F & 0.175 & 0.1891 & 0.1684 & 0.1960 & 0.1520 & \underline{0.1429} & 0.1522 & \textbf{0.1413} & \signcolor{-7.15} \\
					\midrule
					
					\multirow{2}{*}{Azobenzene} 
					& E & 0.025 & 0.0161 & 0.0277 & \underline{0.0161} & \textbf{0.0156} & 0.0394 & 0.0277 & 0.0185 & \signcolor{-33.21} \\					
					& F & 0.097 & 0.0669 & 0.0600 & 0.0761 & \underline{0.0585} & \textbf{0.0513} & 0.0692 & 0.0653 & \signcolor{-5.59} \\
					\midrule
					
					\multirow{2}{*}{Benzene} 
					& E & 0.002 & \underline{0.0009} & 0.0069 & \textbf{0.0007} & \textbf{0.0007} & 0.0096 & 0.0092 & 0.0048 & \signcolor{-47.83} \\					
					& F & 0.017 & 0.0069 & \textbf{0.0046} & 0.0069 & 0.0056 & \underline{0.0047} & 0.0069 & 0.0064 & \signcolor{-7.25} \\
					\midrule
					
					\multirow{2}{*}{Ethanol} 
					& E & 0.014 & 0.0092 & 0.0092 & 0.0092 & 0.0078 & 0.0096 & \underline{0.0032} & \textbf{0.0026} & \signcolor{-18.75} \\					
					& F & 0.085 & 0.0646 & \underline{0.0484} & 0.0738 & 0.0522 & 0.0516 & \underline{0.0484} & \textbf{0.0459} & \signcolor{-5.17}\\
					\midrule
					
					\multirow{2}{*}{Malonaldehyde} 
					& E & 0.025 & 0.0184 & 0.0138 & 0.0185 & \underline{0.0132} & 0.0168 & 0.0185 & \textbf{0.0091} & \signcolor{-50.81}\\					
					& F & 0.152 & 0.1176 & \textbf{0.0830} & 0.1338 & 0.0893 & \underline{0.0875} & 0.0946 & 0.0896 & \signcolor{-5.29} \\
					\midrule
					
					\multirow{2}{*}{Naphthalene} 
					& E & 0.011 & \textbf{0.0046} & \textbf{0.0046} & \textbf{0.0046} & \underline{0.0057} & 0.0174 & 0.0115 & 0.0142 & \signcolor{+23.48} \\					
					& F & 0.060 & 0.0300 & \textbf{0.0208} & 0.0415 & 0.0291 & \underline{0.0242} & 0.0369 & 0.0357 & \signcolor{-3.25} \\
					\midrule
					
					\multirow{2}{*}{Paracetamol} 
					& E & 0.044 & 0.0323 & 0.0346 & 0.0300 & \textbf{0.0258} & 0.0362 & 0.0300 & \underline{0.0264} & \signcolor{-12.00} \\					
					& F & 0.164 & 0.1361 & 0.1130 & 0.1338 & \underline{0.1029} & \textbf{0.0979} & 0.1107 & 0.1054 & \signcolor{-4.76} \\
					\midrule
					
					\multirow{2}{*}{Salicylicacid} 
					& E & 0.017 & \textbf{0.0161} & 0.0208 & 0.0285 & \underline{0.0161} & 0.0330 & 0.0208 & 0.0248 & \signcolor{+19.23} \\					
					& F & 0.088 & 0.0922 & \underline{0.0669} & 0.0992 & 0.0795 & 0.0771 & 0.0715 & \textbf{0.0645} & \signcolor{-9.79} \\
					\midrule
					
					\multirow{2}{*}{Toluene} 
					& E & 0.010 & 0.0069 & 0.0092 & \underline{0.0069} & \textbf{0.0059} & 0.0139 & 0.0115 & 0.0158 & \signcolor{+37.39}\\					
					& F & 0.058 & 0.0369 & 0.0415 & 0.0438 & \underline{0.0264} & \textbf{0.0244} & 0.0350 & 0.0325 & \signcolor{-7.14} \\
					\midrule
					
					\multirow{2}{*}{Uracil} 
					& E & 0.013 & 0.0092 & 0.0138 & \underline{0.0092} & \textbf{0.0069} & 0.0149 & 0.0115 & 0.0146 &  \signcolor{+26.96}\\					
					& F & 0.088 & 0.0669 & \textbf{0.0415} & 0.0738 & 0.0495 & 0.0487 & 0.0484 & \underline{0.0460} &  \signcolor{-4.96} \\

				\midrule
				\multirow{2}{*}{Average} 
                    & E & 0.022 & {0.0175} & 0.0191 & 0.0177 & \textbf{0.0142} & 0.0239 & 0.0195 & \underline{0.0173} & \signcolor{-11.28}\\
                    
                    & F & 0.098 & 0.0807 & 0.0648 & 0.0879 & 0.0645 & \textbf{0.0610} & 0.0674 & \underline{0.0633} & \signcolor{-6.06}  \\
                    					
					\bottomrule
				\end{tabular}
			}
			\end{sc}
		\end{small}
	\end{center}
	\vskip -0.1in
\end{table*}

\begin{table*}[ht]
    \caption{Tail-sensitive force errors (MAE in kcal/mol/\AA) on the rMD17 benchmark, highlighting model robustness for molecular dynamics applications. We report Q$_{95}$, Q$_{99}$, and MAX for both MACE baseline and \maceModulename across all molecules, including the relative differences ($\Delta$) expressed as percentages.}
    \label{table:ablation_Max}
	\begin{center}
		\begin{small}
			\begin{sc}
				\resizebox{0.98\linewidth}{!}{
				\begin{tabular}{rr cccccccccc}
					\toprule
					Model & Metric & Aspirin & Azobenz. & Benzene & Ethanol & Malonal. & Naphtha. & Paracet. & Sal. acid & Toulene & Uracil \\

                    \midrule

					\multirow{3}{*}{Baseline}
                    & Q$_{95}$ & 0.4739 & 0.2327 & 0.0240 & 0.1917 & 0.3454 & 0.1182 & 0.3556 & 0.2314 & 0.1157 & 0.1648 \\
                    
					& Q$_{99}$ & 0.5330 & 0.2661 & 0.0274 & 0.2347 & 0.3978 & 0.1368 & 0.4027 & 0.2657 & 0.1314 & 0.1890 \\
					
                    & MAX & 1.2985 & 0.6496 & 0.0988 & 1.2049 & 1.5461 & \textbf{0.2885} & 1.5136 & 1.1571 & 0.3710 & 0.7688 \\
                    
                    \midrule
					\multirow{3}{*}{\modulenameShort}
                    & Q$_{95}$ & \textbf{0.4393} & \textbf{0.2211} & \textbf{0.0224} & \textbf{0.1805} & \textbf{0.3290} & \textbf{0.1154} & \textbf{0.3341} & \textbf{0.2122} & \textbf{0.1087} & \textbf{0.1593} \\
                    
					& Q$_{99}$ & \textbf{0.5028} & \textbf{0.2526} & \textbf{0.0260} & \textbf{0.2108} & \textbf{0.3867} & \textbf{0.1312} & \textbf{0.3762} & \textbf{0.2397} & \textbf{0.1246} & \textbf{0.1859} \\
					
                    & MAX & \textbf{1.0071} & \textbf{0.5829} & \textbf{0.0808} & \textbf{0.8897} & \textbf{0.9987} & {0.3102} & \textbf{1.2903} & \textbf{1.0989} & \textbf{0.3497} & \textbf{0.6019} \\
                    
                    \midrule
                    
				\multirow{3}{*}{$\Delta (\%)$}
                    & Q$_{95}$ 
                    & \signcolor{-7.31} 
                    & \signcolor{-4.99} 
                    & \signcolor{-6.67} 
                    & \signcolor{-5.84} 
                    & \signcolor{-4.73} 
                    & \signcolor{-2.37} 
                    & \signcolor{-6.12} 
                    & \signcolor{-8.30}
                    & \signcolor{-6.11} 
                    & \signcolor{-3.31}  \\
                    
				& Q$_{99}$ 
                    & \signcolor{-5.67} 
                    & \signcolor{-5.09} 
                    & \signcolor{-5.11} 
                    & \signcolor{-10.21} 
                    & \signcolor{-2.79} 
                    & \signcolor{-4.12} 
                    & \signcolor{-6.57} 
                    & \signcolor{-9.76}
                    & \signcolor{-5.17} 
                    & \signcolor{-1.64}  \\
                    & MAX 
                    & \signcolor{-22.44} 
                    & \signcolor{-10.27} 
                    & \signcolor{-18.18} 
                    & \signcolor{-26.16} 
                    & \signcolor{-35.42} 
                    & \signcolor{+7.52} 
                    & \signcolor{-14.76}
                    & \signcolor{-5.05}
                    & \signcolor{-5.67} 
                    & \signcolor{-21.73} \\
                    
					\bottomrule
				\end{tabular}
                }
			\end{sc}
		\end{small}
	\end{center}
	\vskip -0.2in
\end{table*}

\begin{figure*}[ht]
	\vskip 0.1in
	\begin{center}
		\centerline{\includegraphics[width=0.99\textwidth]{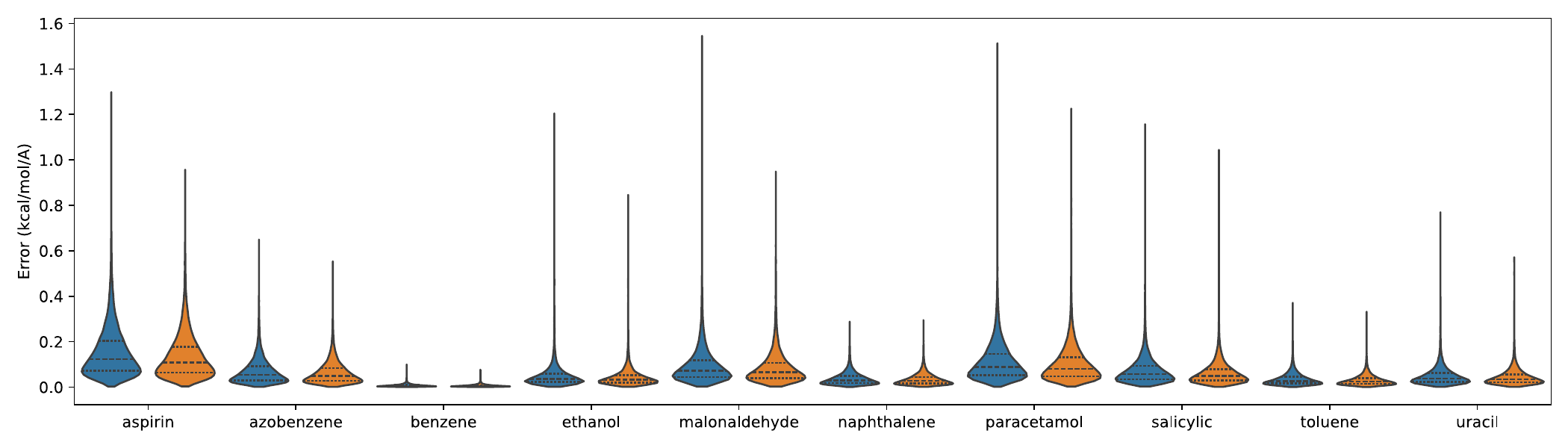}}
		\caption{Violin plots of force error distributions for rMD17 molecules. Blue violins represent the MACE baseline, and orange represent \maceModulename. Each molecule shows separate distributions for the two models, illustrating the overall error spread.}
		\label{fig:distro_errors}
	\end{center}
\end{figure*}

\section{Experiments}

\paragraph{Dataset} We evaluate the proposed method on two established benchmarks for machine learning force fields: revised MD17 (rMD17)~\cite{DBLP:journals/mlst/ChristensenL20} and MD22~\cite{chmiela2023accurate}.
The rMD17 dataset is a reimplementation of MD17~\cite{chmiela2017machine}, which resolves numerical inconsistencies in the original benchmark with higher accuracy. In contrast, MD22 includes larger and more flexible molecules, resulting in more complex potential energy surfaces.

Both datasets consist of \emph{ab initio} molecular dynamics trajectories of organic molecules and provide standardized benchmarks for the development and evaluation of machine learning models of molecular potential energy surfaces.
Model performance is reported in terms of mean absolute error (MAE) for both energies (kcal/mol) and forces (kcal/mol/\AA).

\paragraph{Training} We train the MACE baseline in the high-dimensional configuration and following exactly the training protocol described in the original work~\cite{Batatia2022Design, Batatia2022mace}. We retrain the \maceModulename in the same deterministic environment and identical settings, ensuring that any observed performance differences are solely attributable to the proposed module.

The attention module is implemented as a sinusoidal gating mechanism to modulate the contribution of neighboring atoms based on their positions. We employ the same hyperparameters across all molecules: a $4 \times 8$ grid corresponding to 32 equiangular sampling points, single-head attention, and no attention dropout. All experiments are implemented in PyTorch and trained on an NVIDIA H100 GPU.

\subsection{Results}
Table~\ref{table:ablation_SOTA} shows the MAEs on the rMD17 benchmark, comparing our \maceModulename to several state-of-the-art models, including NequIP~\cite{DBLP:journals/corr/abs-2101-03164}, MACE~\cite{Batatia2022mace},  Allegro~\cite{DBLP:journals/corr/abs-2204-05249}, BOTNet~\cite{Batatia2022Design}, UNiTE~\cite{doi:10.1073/pnas.2205221119}, QuinNet~\cite{DBLP:conf/nips/WangLZWS23}, and VisNet~\cite{wang2024enhancing}.

The introduction of \modulenameShort systematically improves the MACE in force (F) prediction and in the majority of energy (E) prediction tasks (6 out of 10 molecules). Energy MAE reductions of up to 51\% are observed, with an average reduction of about 11\%, while force errors are consistently reduced by 3-10\%, with an average improvement of about 6\%. 

These improvements allow \maceModulename to achieve the lowest MAE for forces for 3 out of 10 molecules, compared to zero for the baseline model. For energy predictions, \modulenameShort allows MACE to achieve the first place for 3 out of 10 molecules, whereas the baseline only reaches a single second-place ranking. These results are obtained on a wide variety of chemical systems, ranging from small aromatic molecules to flexible structures with hydrogen bonds.

Although no model dominates both tasks, MACE enhanced with \modulenameShort ranks second on average in energy and force predictions, outperforming the MACE baseline and matching the performance of the latest state-of-the-art models.

\begin{table*}[ht]
	\caption{Ablation study on the effect of the angular momentum order ($\ell$) in \maceModulename. Energy (E, in kcal/mol) and force (F, in kcal/mol/Å) mean absolute errors are reported for different $\ell$ values across selected molecules, comparing the baseline MACE and the \maceModulename. Relative differences ($\Delta$) indicate the impact of the module at each order.}
	\label{table:ablation_L}
	
	\begin{center}
		\begin{small}
			\begin{sc}
				\resizebox{0.98\linewidth}{!}{
					\begin{tabular}{lcc ccr ccr ccr}
						\toprule
						& & & \multicolumn{3}{c}{$\ell$=1} & \multicolumn{3}{c}{$\ell$=2} & \multicolumn{3}{c}{$\ell$=3} \\
						
						\cmidrule(lr){4-6}
						\cmidrule(lr){7-9}
						\cmidrule(lr){10-12}
						
						Molecule & Atoms &  & MACE & \maceModulename & $\Delta$ (\%)  & MACE & \maceModulename & $\Delta$ (\%)& MACE & \maceModulename & $\Delta$ (\%) \\
						\midrule
						
						\multirow{2}{*}{Ethanol} & \multirow{2}{*}{9}
						& E & 0.0061 & \textbf{0.0045} &  \signcolor{-26.23} & \textbf{0.0031} & 0.0037 &  \signcolor{+19.35} & 0.0032 & \textbf{0.0026} &  \signcolor{-18.75} \\
						&
						& F & 0.0517 & \textbf{0.0474} & \signcolor{-8.32} & 0.0504 & \textbf{0.0458} &  \signcolor{-9.13} & 0.0484 & \textbf{0.0459} &  \signcolor{-5.17} \\
						\midrule
						
						\multirow{2}{*}{Paracetamol} & \multirow{2}{*}{20}
						& E & 0.0916 & \textbf{0.0616} & \signcolor{-32.75} & 0.0347 & \textbf{0.0284} & \signcolor{-18.16} & 0.0300 & \textbf{0.0265} &  \signcolor{-11.67}\\
						&
						& F & 0.1164 & \textbf{0.1134} & \signcolor{-2.58} & 0.1114 & \textbf{0.1072} & \signcolor{-3.71} & 0.1107 & \textbf{0.1054} & \signcolor{-4.77}  \\
						\midrule
						
						\multirow{2}{*}{Ac-Ala3-NHMe} & \multirow{2}{*}{42}
						& E & \textbf{0.0782} & 0.0822 & \signcolor{+5.13} & 0.0871 & \textbf{0.0821} & \signcolor{-5.74} & 0.0883 & \textbf{0.0803} & \signcolor{-9.02} \\
						&
						& F & 0.1322 & \textbf{0.1319} & \signcolor{-0.23} & 0.1219 & \textbf{0.1197} & \signcolor{-1.81} & 0.1230 & \textbf{0.1213} & \signcolor{-1.30} \\		
						
						\midrule
						
						\multirow{2}{*}{Stachyose} & \multirow{2}{*}{87} 
						& E & {0.1223} & 0.1223 & \signcolor{0.00} & \textbf{0.1044} & {0.1154} & \signcolor{+10.54} & 0.1325 & \textbf{0.1074} & \signcolor{-18.93} \\
						& 
						& F & {0.1791} & \textbf{0.1724} & \signcolor{-3.74} & 0.1635 & \textbf{0.1534} & \signcolor{-6.16} & 0.1634 & \textbf{0.1524} & \signcolor{-6.77} \\
						
						\bottomrule
					\end{tabular}
				}
			\end{sc}
		\end{small}
	\end{center}
	\vskip -0.1in
\end{table*}

\subsection{Tail-risk and error distribution analysis}

While the MAE is the standard metric for evaluating machine learning force fields, previous studies have shown that it does not fully capture model behavior in molecular dynamics settings~\cite{DBLP:journals/tmlr/0005WWXKGJ23, DUANGDANGCHOTE20242177}. In these contexts, beyond average force accuracy, controlling the distribution of errors is crucial, since the accumulation of rare but large errors can drive the system toward unphysical configurations and compromise simulation stability~\cite{10.1063/5.0147023, DBLP:journals/corr/abs-2506-14850}.

To investigate this, we analyze error distributions both qualitatively and quantitatively, using tail-sensitive metrics: the 95th percentile error (Q$_{95}$), the 99th percentile error (Q$_{99}$), and the maximum observed error (MAX). Table~\ref{table:ablation_Max} reports these metrics for \maceModulename and the MACE baseline across all molecules in the rMD17 dataset.

The introduction of \modulenameShort consistently improves Q$_{95}$ and Q$_{99}$ across all systems, with average reductions of 5.58\% and 5.51\%, respectively, indicating a systematic mitigation of high-error events. Improvements are also observed for the maximum error in most molecules, with an average reduction of 15.12\%. A single exception occurs for naphthalene, where the maximum error increases by 7.52\%; however, this effect is confined to the extreme tail and does not affect Q$_{95}$ or Q$_{99}$, which remain improved.

Overall, these results suggest that \modulenameShort reduces both tail and worst-case errors, complementing the improvements in mean accuracy. This reduction in extreme errors lowers the risk that rare, high-magnitude deviations adversely impact molecular dynamics simulations, thereby enhancing model robustness in practical applications. To assess the practical impact of MARA, we conduct a molecular dynamics (MD) simulation test (details in Appendix~\ref{label:MD_traj}).

\subsection{Ablations}

We conduct ablation studies to analyze the impact of key hyperparameters of \maceModulename on performance. All experiments are performed in a deterministic setting, keeping all conditions fixed except for the parameter under investigation. We consider two orthogonal experiments.

\paragraph{Effect of the harmonic order $\ell$} First, we vary the harmonic order $\ell$, which determines the angular resolution of the model. Larger values of $\ell$ enable the representation of finer geometric details at the cost of increased computational complexity, while smaller values reduce angular resolution. Table~\ref{table:ablation_L} reports energy (E) and force (F) mean absolute errors for selected molecules, comparing the MACE baseline with \maceModulename.

Across the rMD17 and MD22 datasets, \modulenameShort, using a $4 \times 8$ grid, consistently improves force predictions and generally enhances energy predictions. Larger and more flexible molecules tend to benefit more from higher angular resolution, whereas lower $\ell$ values are sufficient for smaller and more rigid molecules. These trends indicate that, by varying $\ell$, the module continues to effectively capture angular dependencies across molecules of different sizes and complexities, providing a robust and scalable improvement over the baseline.

\begin{table}[t]
	\caption{Ablation study on the effect of grid resolution for \maceModulename ($\ell=2$) on Paracetamol. Energy (E, kcal/mol) and force (F, kcal/mol/Å) MAEs are reported, along with inference time per batch (ms). Percentage differences ($\Delta\%$) are relative to the baseline Paracetamol $\ell=2$ model.}
    \label{table:ablation_grid}
	\begin{center}
		\begin{small}
			\begin{sc}
				\resizebox{0.98\linewidth}{!}{
				\begin{tabular}{rccccc}
					\toprule
                    & \multicolumn{5}{c}{Grid Size} \\
                    \cmidrule (lr){2-6}
                    & $2 \times 2$ & $2 \times 4$ & $4 \times 4$ & $4 \times 8$ & $8 \times 8$ \\
                    \midrule
                    N. Points & 4 & 8 & 16 & 32 & 64\\
                    \midrule
                    Energy & 0.0332 & 0.0411 & 0.0379 & \textbf{0.0284} & 0.0300 \\
                    \midrule
                    $\Delta$ Energy (\%)& \signcolor{-4.33}  & \signcolor{+18.44}  & \signcolor{+9.22}  & \signcolor{-18.16}  & \signcolor{-13.54}  \\
                    \midrule
                    Force & 0.1089 & 0.1057 & 0.1075 & 0.1072 & \textbf{0.1045} \\
                    \midrule
                    $\Delta$  Force (\%)& \signcolor{-2.24}  & \signcolor{-5.13}  & \signcolor{-3.43}  & \signcolor{-3.73}  & \signcolor{-6.21} \\
                    \midrule
                    Time (ms) & \textbf{111.67} & 112.74 & 114.15 & 117.97 & 128.18 \\
                    \midrule
                    $\Delta$ Time (\%)& \textbf{+9.37} & +10.42 & +11.80 & +15.54 & +25.54 \\
					\bottomrule
				\end{tabular}
                }
			\end{sc}
		\end{small}
	\end{center}
	\vskip -0.1in
\end{table}

\begin{figure*}[h]
    \centering
    \begin{subfigure}{0.46\textwidth}
        \centering
        \includegraphics[width=\linewidth]{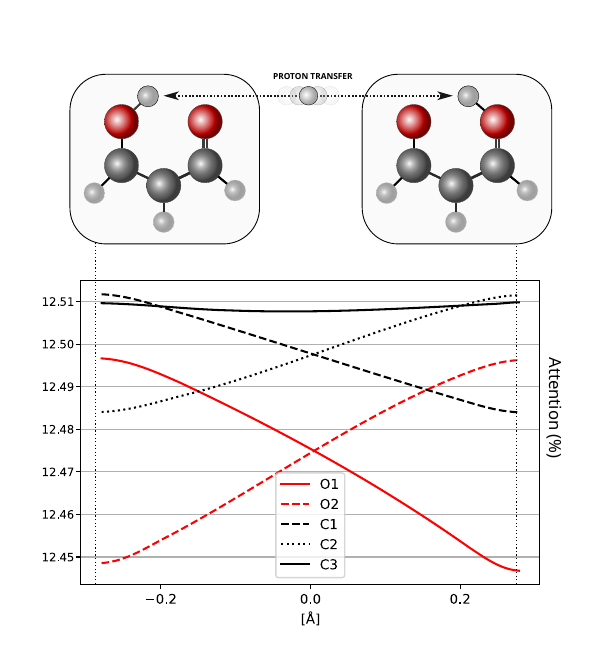}
        \caption{Attention received by the $H_1$ atom}
        \label{fig:attention_r3}
    \end{subfigure}
    \hfill
    \begin{subfigure}{0.46\textwidth}
        \centering
        \includegraphics[width=\linewidth]{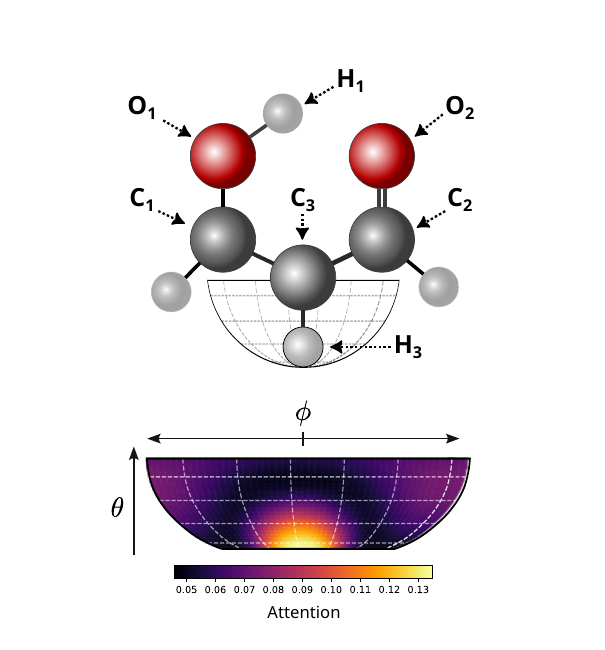}
        \caption{Attention Map between $C_3$ and $H_3$}
        \label{fig:attention_s2}
    \end{subfigure}
    \caption{Interpretability of MARA. (a) Radial slice of the attention map showing attention from heavy atoms ($O_1$, $O_2$, $C_1$, $C_2$, $C_3$) to hydrogen ($H_1$) during intramolecular proton transfer. (b) Angular slice of the attention map showing attention between $C_3$ and $H_3$ as $H_3$ is rotated around $C_3$ at a fixed distance of $1,\text{\AA}$}
    \label{fig:main}
\end{figure*}

\paragraph{Effect of grid resolution} Second, we fix $\ell=2$ and vary the grid resolution for Paracetamol (Table~\ref{table:ablation_grid}). Percentage differences are shown relative to the baseline Paracetamol $\ell=2$ model (Table~\ref{table:ablation_L}). Increasing the number of grid points improves both energy and force predictions, with the $4 \times 8$ and $8 \times 8$ grids achieving the largest gains. As expected, finer grids also increase computational cost: inference wall-clock time per batch rises by +9.4\% for a $2 \times 2$ grid and +25\% for $8 \times 8$ relative to the baseline, measured on an NVIDIA RTX 4090, with an average increase of 14.5\%. These results highlight the trade-off between accuracy and efficiency and confirm that \modulenameShort enhances performance across grid configurations.

Taken together, these ablation experiments demonstrate that \modulenameShort provides robust improvements across both angular and radial resolutions, enhancing force and energy predictions consistently while balancing accuracy and computational costs depending on molecular size and complexity.

\subsection{Interpretability of attention maps}

In this section, we analyze the model through attention map analysis, using malonaldehyde as a case study. Two experiments are conducted to explore the effect of \modulenameShort's attention on atomic movement.

In the first experiment, we study the intramolecular proton transfer in malonaldehyde, a process involving one hydrogen atom and two oxygen atoms~\cite{Schroder2011}. This transfer occurs through a proton hopping mechanism that leverages hydrogen bonds and quantum tunneling phenomena, with a timescale on the order of femtoseconds~\cite{zhou2025state}. We simulate the process by moving the hydrogen atom $H_1$ between the two oxygen atoms in 100 steps, recording the attention that the hydrogen atom receives from the surrounding oxygen ($O_1$, $O_2$) and carbon ($C_1$, $C_2$, $C_3$) atoms. The results, shown in Figure~\ref{fig:attention_r3}, reveal that the radial slice of the attention map is symmetric around the saddle point, with carbon atoms exerting more attention on hydrogen compared to the oxygen atoms. Additionally, the intensity of attention varies as a function of the distance between the atoms.

In the second experiment, we examine attention in relation to angular geometry. The distance between $C_3$ and $H_3$ is fixed at $1 \AA$, and the hydrogen atom is rotated around carbon $C_3$, forming a hemisphere. Attention between the two atoms is monitored at each rotation angle. The angular slices of the attention map, shown in Figure~\ref{fig:attention_s2}, reveal that attention intensity is maximized at a specific angle between the two atoms and varies significantly with the angle. This confirms that attention depends not only on distance but also on angular geometry, consistent with findings from the chemical literature~\cite{doi:10.1021/acs.jpca.9b10248, doi:10.1021/acs.jctc.3c00907}.

These results suggest that \modulenameShort, in addition to improving and stabilizing the performance of SE(3)-equivariant models, could be valuable in computational chemistry for analyzing complex angular and radial patterns in molecular interactions.

\section{Conclusion}
In this work, we introduce \modulenameFull~(\modulenameShort), a continuous spherical attention mechanism that provides an efficient approximation to SE(3)-equivariant interactions in MLFF models.~\modulenameShort operates directly on the angular and radial coordinates of atomic neighbors, enabling flexible and modular geometric weighting of local interactions.

Its plug-and-play integration, as demonstrated for MACE, requires no modifications to the underlying model, making it suitable for other SE(3)-equivariant architectures. The method consistently improves accuracy and robustness in the prediction of forces, while also reducing high-error cases. These results suggest that \modulenameShort is an effective and generalizable operator capable of enhancing the expressiveness of atomistic models, with potential impact on the stability and reliability of atomistic simulations.

\paragraph{Limitations and future work,}
\modulenameShort provides an efficient approximation to SE(3)-equivariant interactions via spherical discretization and local neighborhood cutoffs. This introduces tunable parameters such as grid resolution, enabling a controllable trade-off between accuracy and computational cost depending on the complexity of the system. Our evaluation focuses on molecular benchmarks using a single host architecture as the primary backbone. While we demonstrate the portability of \modulenameShort by integrating it into other SE(3)-equivariant models (see Appendix~\ref{app:othersplug}), broader validation on diverse chemical systems and different architectural paradigms remains future work.

\section*{Impact Statement}

This paper presents work whose goal is to advance the field of Machine
Learning. There are many potential societal consequences of our work, none
which we feel must be specifically highlighted here.

\section*{Acknowledgment}
This study is funded by the Swiss National Science Foundation (SNSF) - Project Nr 200020 219388.

\nocite{langley00}
\bibliography{ICML26_bib.bib}
\bibliographystyle{icml2026}

\newpage
\appendix
\onecolumn





\section{Mathematical Background}

In this section, we provide the mathematical background for the constructions used in the main text. First, we discuss the basic properties of the two-dimensional sphere $S^2$ and the special orthogonal group $SO(3)$, which governs rotations in three-dimensional space.
We then describe the special Euclidean group $SE(3)$, which represents the motions of rigid bodies, and finally introduce the grid field used in our algorithm along with its key properties, including scale linearity and invariance under rigid transformations.

\subsection{Rotations on the 2-Sphere}

We discuss some fundamental notions concerning the two-dimensional sphere $S^2$ and the rotation group $SO(3)$.

 To define functions on the sphere, we require coordinates. A standard choice is to parametrize points \( \mathbf {p} \in S^2 \) using the colatitude \( \theta \in [0, \pi] \) and the longitude \( \varphi \in [0, 2\pi] \). In these coordinates, a unit vector \( \mathbf{p} \) can be expressed as
 \begin{equation}
    \mathbf{p} = (\cos(\varphi) \sin(\theta), \sin(\varphi) \sin(\theta), \cos(\theta))^T.
\end{equation}


To discuss symmetries of the sphere, we consider the special orthogonal group \(SO(3)\), which consists of all proper rotations in \(\mathbb{R}^3\). Elements of \(SO(3)\) are \(3\times 3\) matrices with determinant one, whose inverses coincide with their transposes. Any rotation \( R \in SO(3) \) can be written in terms of the Eulerian angles $\varphi \in [0, 2 \pi], \theta \in [0, \pi], \psi \in [0, 2 \pi]$, such that

\begin{equation}
    R = R_z(\varphi) R_y(\theta) R_z(\psi),
\end{equation}
where $R_z$ and $R_y$ are rotations around the z- and y-axes:
\begin{equation}
    R_z(\varphi) = \mqty[\cos \varphi & - \sin \varphi & 0 \\ \sin \varphi & \cos \varphi & 0 \\ 0 & 0 & 1], \quad
    R_y(\varphi) = \mqty[\cos \varphi  & 0 & \sin \varphi\\ 0 & 1 & 0 \\ -\sin \varphi & 0 & \cos \varphi ].
\end{equation}
Unlike translations in the plane, rotations generally do not commute, making $SO(3)$ non-abelian. Rotations in \(SO(3)\) can reach any point on the sphere. In particular, applying a rotation to the north pole \( n = (0, 0, 1)^T \) yields

\begin{equation}
    Rn = R_z(\varphi) R_y(\theta) R_z(\psi) n = (\cos(\varphi) \sin(\theta), \sin(\varphi) \sin(\theta), \cos(\theta))^T. 
\end{equation}
We observe that the last rotation angle $\psi$ is dropped, illustrating that $S^2$ can be obtained as the quotient of $SO(3)$ and $SO(2)$.

\subsection{Rigid motions and the special euclidean group}

The special euclidean group \(SE(3)\) describes all rigid body motions in three-dimensional space, combining rotations and translations. Each element of \(SE(3)\) can be represented as a pair \((R, \mathbf{t})\), with \(R \in SO(3)\) and \(\mathbf{t} \in \mathbb{R}^3\), or equivalently as a \(4\times 4\) homogeneous transformation matrix:
\begin{equation}
    T =
    \begin{bmatrix}
        R & \mathbf{t} \\
        \mathbf{0}^\mathrm{T} & 1
    \end{bmatrix}.
\end{equation}

Composition of two rigid motions corresponds to matrix multiplication:
\begin{equation}
    T_1 T_2 =
    \begin{bmatrix}
    R_1 R_2 & R_1 \mathbf{t}_2 + \mathbf{t}_1 \\
    \mathbf{0}^\mathrm{T} & 1
    \end{bmatrix}.
\end{equation}

Unlike \(SO(3)\), \(SE(3)\) also includes translations. It is a six-dimensional, non-abelian Lie group: three degrees of freedom correspond to rotation, and three to translation. The subgroup of pure translations,
\begin{equation}
    \mathbb{R}^3 \cong \{ (\mathbb{I}_3, \mathbf{t}) \mid \mathbf{t} \in \mathbb{R}^3 \},
\end{equation}
is normal in \(SE(3)\), and the group can be expressed as a semi-direct product:
\begin{equation}
    SE(3) \cong \mathbb{R}^3 \rtimes SO(3).
\end{equation}

Rigid motions in \(SE(3)\) are fundamental in robotics, computer vision, and physics, as they describe the complete spatial displacement of a body in \(\mathbb{R}^3\). As a smooth Lie group, \(SE(3)\) generalizes the rotational symmetries of \(SO(3)\) to full rigid body transformations.

\subsection{Properties of the grid field}

Consider two points $\mathbf{p}_S, \mathbf{p}_T \in \mathbb{R}^3$ and define the vector connecting them:
\begin{equation}
    \mathbf{v} = \mathbf{p}_S - \mathbf{p}_T, \quad d = \|\mathbf{v}\|, \quad \hat{\mathbf{u}} = \frac{\mathbf{v}}{d} \in S^2.
\end{equation}

Let
\begin{equation}
    \{ g_k \in \mathbb{R}^3 : \|g_k\| = 1, \, k = 1, \dots, N \}
\end{equation}

denote a set of points on the unit sphere $S^2$. We define a scaled sphere of radius $d$ centered at $\mathbf{p}_S$:
\begin{equation}
    \mathbf{x}_k = \mathbf{p}_S + d \, g_k.
\end{equation}

The distance from $\mathbf{x}_k$ to $\mathbf{p}_T$ is
\begin{equation}
    \delta_k = \|\mathbf{x}_k - \mathbf{p}_T\| = \| (\mathbf{p}_S - \mathbf{p}_T) + d g_k \| = \| \mathbf{v} + d g_k \|.
\end{equation}

Dividing by $d$ and using the unit vector $\hat{\mathbf{u}} = \frac{\mathbf{v}}{d}$ gives a normalized distance:
\begin{equation}
    \frac{\delta_k}{d} = \| \hat{\mathbf{u}} + g_k \|.
\end{equation}

Expanding the norm yields
\begin{equation}
    \delta_k = d \, \|\hat{\mathbf{u}} + g_k\| = d \sqrt{ (\hat{\mathbf{u}} + g_k) \cdot (\hat{\mathbf{u}} + g_k) } = d \sqrt{ 2 + 2 (\hat{\mathbf{u}} \cdot g_k) }.
\end{equation}
    
Let $\theta_k$ denote the angle between $\hat{\mathbf{u}}$ and $g_k$, so that $\hat{\mathbf{u}} \cdot g_k = \cos \theta_k$. Then

\begin{equation}
    \delta_k = d \sqrt{ 2 (1 + \cos \theta_k) }.
\end{equation}

In particular, the extreme cases are
\begin{align}
    \begin{cases}
        \cos \theta_k = 1  & \implies \delta_k = 2d \\ 
        \cos \theta_k = -1 & \implies  \delta_k = 0         
    \end{cases}
\end{align}

This shows that $\delta_k$ is linearly proportional to $d$ for any $\theta_k$. Under a rigid transformation
\begin{equation}
    \mathbf{p}'_S = R \mathbf{p}_S + t, \quad \mathbf{p}'_T = R \mathbf{p}_T + t, \quad R \in SO(3), \, t \in \mathbb{R}^3,
\end{equation}

the connecting vector transforms as
\begin{equation}
    \mathbf{v}' = \mathbf{p}'_S - \mathbf{p}'_T = R \mathbf{v}.
\end{equation}

Simultaneously, the unit sphere points are rotated: $g'_k = R g_k$. Therefore, the distances $\delta_k$ are invariant under rigid transformations:
\begin{equation}
    \delta'_k = \| \mathbf{v}' + d g'_k \| = \| R (\mathbf{v} + d g_k) \| = \| \mathbf{v} + d g_k \| = \delta_k.
\end{equation}

Thus, the field defined by $\delta_k$ is linearly sensitive to the scale $d$ and invariant under rigid body transformations.

\section{Architecture}

\begin{figure}[ht]
	\vskip 0.2in
	\begin{center}
		\centerline{\includegraphics[width=\linewidth]{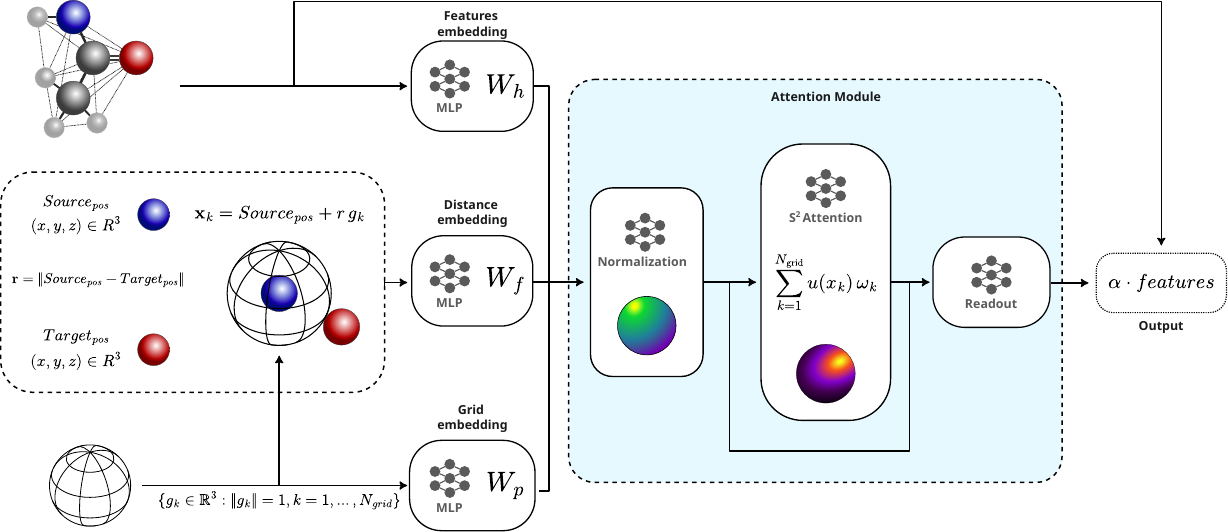}}
		\caption{Module operating diagram}
		\label{img:architectural_sketch}
	\end{center}
\end{figure}

For the implementation of our module, Python 3.10 with support for CUDA 12.8 and Torch 2.8.0 is used. For the spherical attention, the original implementation of~\citet{attention_on_sphere} in the torch-harmonics 0.8.0 package, \href{https://github.com/NVIDIA/torch-harmonics}{available online}, is used~\cite{bonev2023spherical}. The version of ACE/MACE suite~\cite{Batatia2022mace, Batatia2022Design} we use is the one in which the module is introduced in its \href{https://github.com/ACEsuit/mace}{original version}. In addition, the \href{https://github.com/NVIDIA/cuEquivariance}{cuEquivariance library} has been used, which allows optimisations for CUDA on Torch.

The module is used with a $4\times8$ grid, with the exception of Table~\ref{table:ablation_grid}, for which we have modified the resolution to verify its impact. In MACE, the module is introduced in the RealAgnosticInteractionBlock and RealAgnosticResidualInteractionBlock blocks, as shown in Algorithm~\ref{alg:MaceBlock}. The module returns the message already enriched by attention and separately the applied weights. Our module is publicly available at \href{https://github.com/monsieursolver/MARA}{https://github.com/monsieursolver/MARA} for further details and implementations. The model's functioning diagram is shown in the Figure~\ref{img:architectural_sketch}.

\begin{algorithm}[ht]
    \caption{Insertion of the module into the MACE blocks}
    \label{alg:MaceBlock}
    \begin{algorithmic}[1]
        \STATE $m_{ji} \gets \text{self.conv}_{tp}(node_{feats}[edge_{index}[0]],\; edge_{attrs},\; tp_{weights})$
        \STATE $m_{ji},\; att \gets \text{self.spherical\_attention}(m_{ji},\; positions,\; edge_{index},\; edge_{feats})$
        \STATE $message \gets scatter\_sum(src=m_{ji},\; index=edge_{index}[1],\; dim=0,\; dim\_size=node_{feats}.shape[0])$
    \end{algorithmic}
\end{algorithm}

The model has been trained primarily on an NVIDIA H100, maintaining the same parameters used in the original paper, which uses an NVIDIA A100. Inference tests, on the other hand, have been performed using an RTX 4090.

\section{Learnable positional encoding}

We conduct a set new of ablation experiments to assess the contribution of learnable projections and positional encoding to the prediction of energies (E) and forces (F). We select four molecules with substantially different system sizes, fix $\ell = 2$, and use a $4 \times 8$ grid. For each molecule, we retrain the \maceModulename under four configurations: (i) the full model, (ii) without positional encoding, (iii) with fixed (non-learnable) projections, and (iv) without positional encoding and with fixed projections.

\begin{table}[H]
	\caption{Ablation study on the effect of positional encoding and learnable projections for energy (E, MAE kcal/mol)and force (F, MAE kcal/mol/\AA) prediction. We compare the \maceModulename with variants that remove positional encoding, disable learning of the projections, or apply both modifications. Lower values indicate better performance.}
	\label{table:ablation_grid_pos}
	\begin{center}
		\begin{small}
			\begin{sc}
				\begin{tabular}{ccc c cccc}
					\toprule
					\multirow{2}{*}{Learnable} & Not & Positional &
					& \multirow{2}{*}{Ethanol} 
					& \multirow{2}{*}{Paracetamol}
					& \multirow{2}{*}{Ac-Ala3-NHMe}
					& \multirow{2}{*}{Stachyose} \\
					& Learnable & Encoding & & & & & \\					
					\midrule
					
					\multirow{2}{*}{\checkmark} & & \multirow{2}{*}{\checkmark} & 
					E & 0.0037 & \bfseries 0.0284 & 0.0821 & 0.1154 \\
					&&&
					F & \bfseries 0.0458 & \bfseries 0.1072 & 0.1197 & \bfseries 0.1534 \\
					\midrule
					
					& \multirow{2}{*}{\checkmark} & \multirow{2}{*}{\checkmark} & 
					E & 0.0036 & 0.0292 & 0.0745 & 0.1242 \\
					&&&
					F & 0.0487 & 0.1084 & 0.1161 & 0.1564 \\
					\midrule

					\multirow{2}{*}{\checkmark} & & \multirow{2}{*}{} & 
					E & \bfseries 0.0028 & 0.0292 & 0.0728 & 0.1594 \\
					&&&
					F & 0.0486 & 0.1081 & \bfseries 0.1152 & 0.1557 \\
					\midrule
					
					& \multirow{2}{*}{\checkmark} & & 
					E & 0.0036 & 0.0307 & \bfseries 0.0711 & \bfseries 0.1132 \\
					&&&
					F & 0.0484 & 0.1125 & 0.1157 & 0.1555 \\
					
					\bottomrule					
					
				\end{tabular}
			\end{sc}
		\end{small}
	\end{center}
	\vskip -0.1in
\end{table}

The results, shown in Table ~\ref{table:ablation_grid_pos}, show that positional encoding and learnable projections generally improve force prediction performance, achieving lower mean absolute error on three of the four molecules considered. In contrast, for energy prediction, only one molecule achieves its minimum error in the same configuration. The best overall energy performance (two out of four molecules) is achieved with the opposite setting, i.e., without positional encoding and without learnable projections.

Finally, these results indicate that retaining positional encoding while disabling learning of the projections is not beneficial: this configuration does not achieve the best performance in any of the eight evaluated metrics.

\section{Convergence study}

\begin{table}[H]
	\caption{Test-set performance on MD22 DHA and AT–AT DNA base pairs for different training regimes. We compare MACE baseline and \maceModulename with 100 and 10,000 samples, reporting errors on energies (E, MAE in kcal/mol) and forces (F, MAE in kcal/mol/\AA).}
	\label{table:ablation_size}
	\begin{center}
		\begin{small}
			\begin{sc}
				\begin{tabular}{cc ccr ccr}
					\toprule
					
					\multirow{2}{*}{Molecules} & &
					\multicolumn{3}{c}{Training samples = 100} &
					\multicolumn{3}{c}{Training samples = 10K}
					\\
					\cmidrule(lr){2-5}
					\cmidrule(lr){6-8}
					& &
					Baseline & \modulenameShort & $\Delta$ (\%) &
					Baseline & \modulenameShort & $\Delta$ (\%) 	\\		
					\midrule
					
					\multirow{2}{*}{DHA} & 
					E & 0.2685 & \bfseries 0.2674 & \signcolor{-0.41} & 0.1212 & \bfseries 0.1204 & \signcolor{-0.66} \\
					&
					F & 0.4924 & \bfseries 0.4796 & \signcolor{-2.60} & 0.1245 & \bfseries 0.1184 & \signcolor{-4.90} \\
					\midrule
					
					\multirow{2}{*}{AT-AT} & 
					E & \bfseries 0.1324 & 0.1511 & \signcolor{+14.12} & 0.1482 & \bfseries 0.1480 & \signcolor{-0.13} \\
					&
					F & \bfseries 0.1759 & 0.1783 & \signcolor{+1.36} & 0.1274 & \bfseries 0.1209 & \signcolor{-5.10} \\
					
					\bottomrule

				\end{tabular}
			\end{sc}
		\end{small}
	\end{center}
	\vskip -0.1in
\end{table}

We further analyze two additional molecules from the MD22 dataset, specifically Docosahexaenoic Acid (DHA), consisting of 56 atoms, and AT–AT DNA base pairs, consisting of 60 atoms, in order to evaluate the performance of our model under conditions of data scarcity and abundance. The dataset comprises 69,744 and 19,999 molecular dynamics (MD) trajectories, respectively. From each set, we randomly sample 100 and 10,000 configurations for the training set, keeping 1,000 examples for validation and using all remaining configurations as the test set.

As shown in Table~\ref{table:ablation_size}, performance -- particularly in terms of force prediction -- differs significantly between training with 100 examples and training with 10,000 examples. This behavior is expected: with a limited number of samples, the model has limited ability to learn less frequent configurations, leading to less accurate predictions. Furthermore, both molecules considered are highly flexible and have a large conformational space, making the problem particularly challenging in a low-data regime.

The results show that, with few training examples, the proposed module does not provide significant benefits and, in one case, even worsens the prediction of forces compared to the baseline model. On the other hand, when the number of samples is sufficiently high, the attention mechanism is particularly effective, improving performance in both tasks: on average, there is an improvement of about 5\% in forces and about 0.4\% in energies, the latter being considered a marginal gain. This pattern is further supported by Table~\ref{table:ablation_size_max}, which analyzes the distributions of Q${95}$, Q${99}$, and MAX.

\begin{table}[H]
	\caption{Test-set performance on MD22 DHA and AT–AT DNA base pairs for different training regimes. We compare MACE baseline and \maceModulename with 100 and 10,000 samples, reporting errors Q$_{95}$, Q$_{99}$ and max on forces (F, MAE in kcal/mol/\AA).}
	\label{table:ablation_size_max}
	\begin{center}
		\begin{small}
			\begin{sc}
				\begin{tabular}{cc ccr ccr}
					\toprule
					
					\multirow{2}{*}{Molecules} & &
					\multicolumn{3}{c}{Training samples = 100} &
					\multicolumn{3}{c}{Training samples = 10K}
					\\
					\cmidrule(lr){2-5}
					\cmidrule(lr){6-8}
					& &
					Baseline & \modulenameShort & $\Delta$ (\%) &
					Baseline & \modulenameShort & $\Delta$ (\%) 	\\		
					\midrule
					
					\multirow{3}{*}{DHA} & 
					Q$_{95}$ & 1.5110 & \bfseries 1.4893 & \signcolor{-1.44} & 0.3699 & \bfseries 0.3511 & \signcolor{-5.11} \\
					& Q$_{99}$ & 2.5615 & \bfseries 2.5436 & \signcolor{-0.70} & 0.5936 & \bfseries 0.5655 & \signcolor{-4.73} \\
					& MAX & \bfseries 42.857 & 44.344 & \signcolor{+3.47} & \bfseries 5.2803 & 5.8291 & \signcolor{+10.41} \\
					
					\midrule
					
					\multirow{3}{*}{AT-AT} & 
					Q$_{95}$ & \bfseries 0.5385 & 0.5504 & \signcolor{+2.21} & 0.3617 & \bfseries 0.3423 & \signcolor{-5.37} \\
					& Q$_{99}$ & \bfseries 0.9075 & 0.9250 & \signcolor{+1.93} & 0.5483 & \bfseries 0.5177 & \signcolor{-5.59} \\
					& MAX & 18.937 & \bfseries 18.678 & \signcolor{-1.36} & 6.1684 & \bfseries 5.7134 & \signcolor{-7.39} \\
					
					\bottomrule

				\end{tabular}
			\end{sc}
		\end{small}
	\end{center}
	\vskip -0.1in
\end{table}

\subsection{Convergence speed}

To verify that attention provides a real benefit to the training process, we analyze the model's performance at each epoch on the validation set of 1000 examples, displaying both the loss and the corresponding RMSE on both forces and energies. In both configurations, the models are trained for 100k gradient steps, allowing them to reach a condition of near convergence.

To quantify the benefit of introducing \modulenameShort, we also display the relative ratio between the MACE-Baseline model and \maceModulename, defined as:
\[\frac{\text{MACE-Baseline}-\text{\maceModulename}}{\text{MACE-Baseline}} \cdot 100.\]

In this case, positive values indicate that \maceModulename performs better than MACE-Baseline, while negative values indicate that MACE-Baseline is superior, as it has an error closer to zero.

\begin{figure}[H]
	\vskip 0.2in
	\begin{center}
		\centerline{\includegraphics[width=0.9\linewidth]{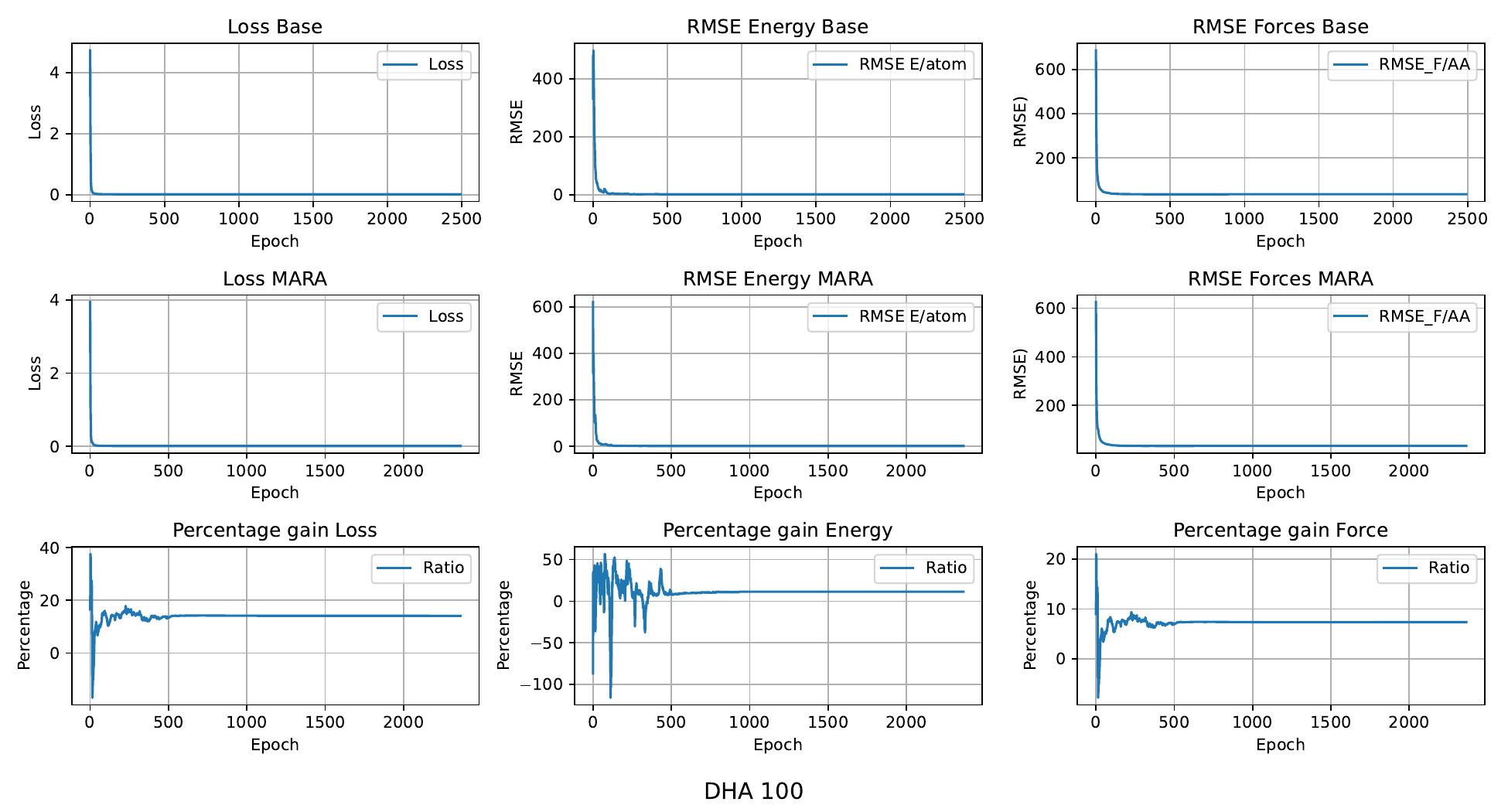}}
		\caption{Performance of models on the validation set on DHA with 100 samples. The subplots show, from left to right, Loss, Energy, and Strength; from top to bottom, Baseline, \modulenameShort, and the percentage ratio.}
		\label{img:DHA100}
	\end{center}
\end{figure}

\begin{figure}[H]
	\begin{center}
		\centerline{\includegraphics[width=0.9\linewidth]{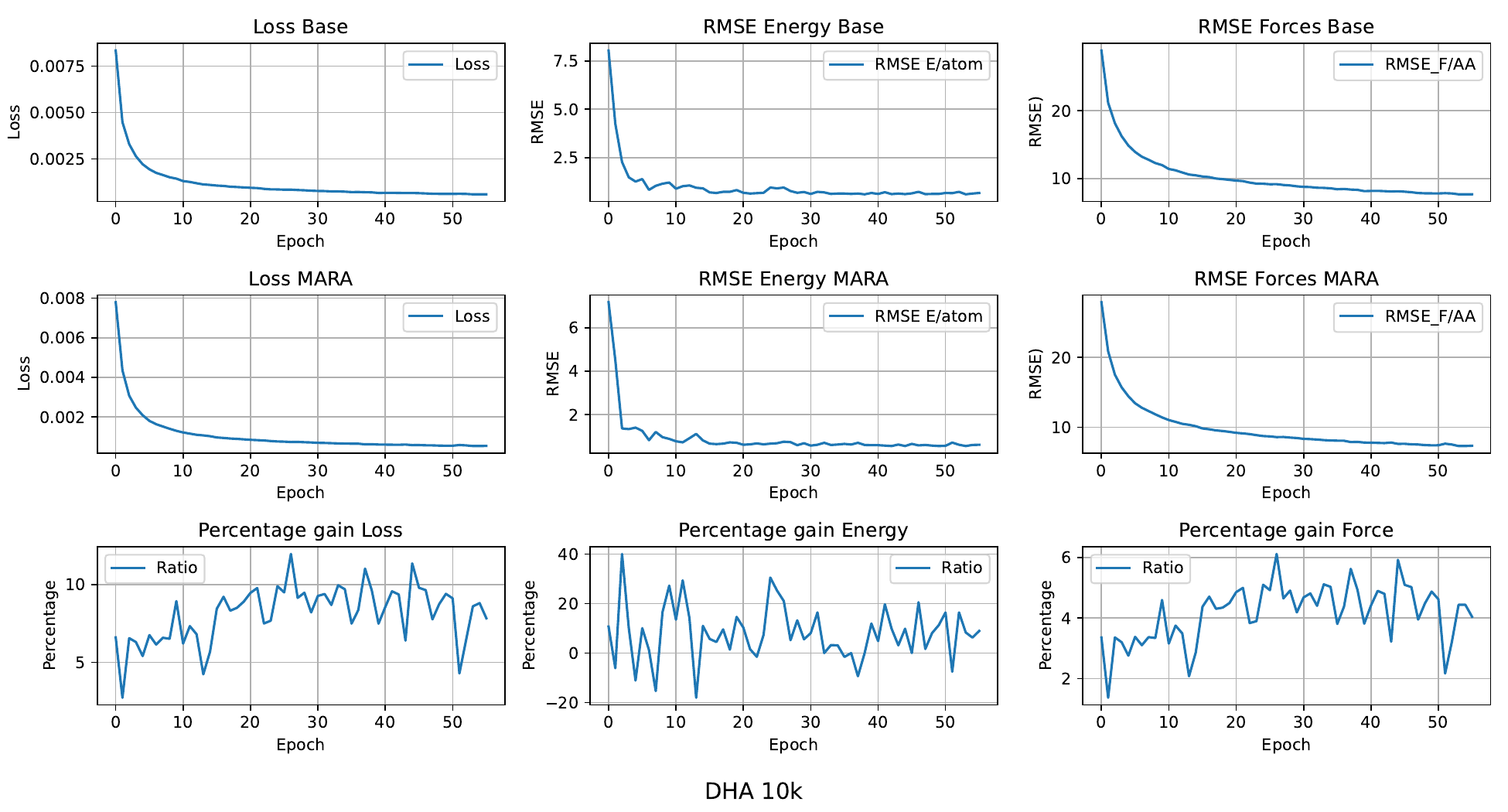}}
		\caption{Performance of models on the validation set on DHA with 10,000 samples. The subplots show, from left to right, Loss, Energy, and Strength; from top to bottom, Baseline, \modulenameShort, and the percentage ratio.}
		\label{img:DHA10000}
	\end{center}
\end{figure}

Figures \ref{img:DHA100} and \ref{img:DHA10000} show that, for Docosahexaenoic Acid (DHA), the validation ratio is positive in both configurations, leading to an advantage in the testing phase. Notably, as training examples increase, \modulenameShort maintains a positive ratio, especially in force prediction, benefiting from early epochs and progressively amplifying the advantage.

\begin{figure}[H]
	\vskip 0.2in
	\begin{center}
		\centerline{\includegraphics[width=0.9\linewidth]{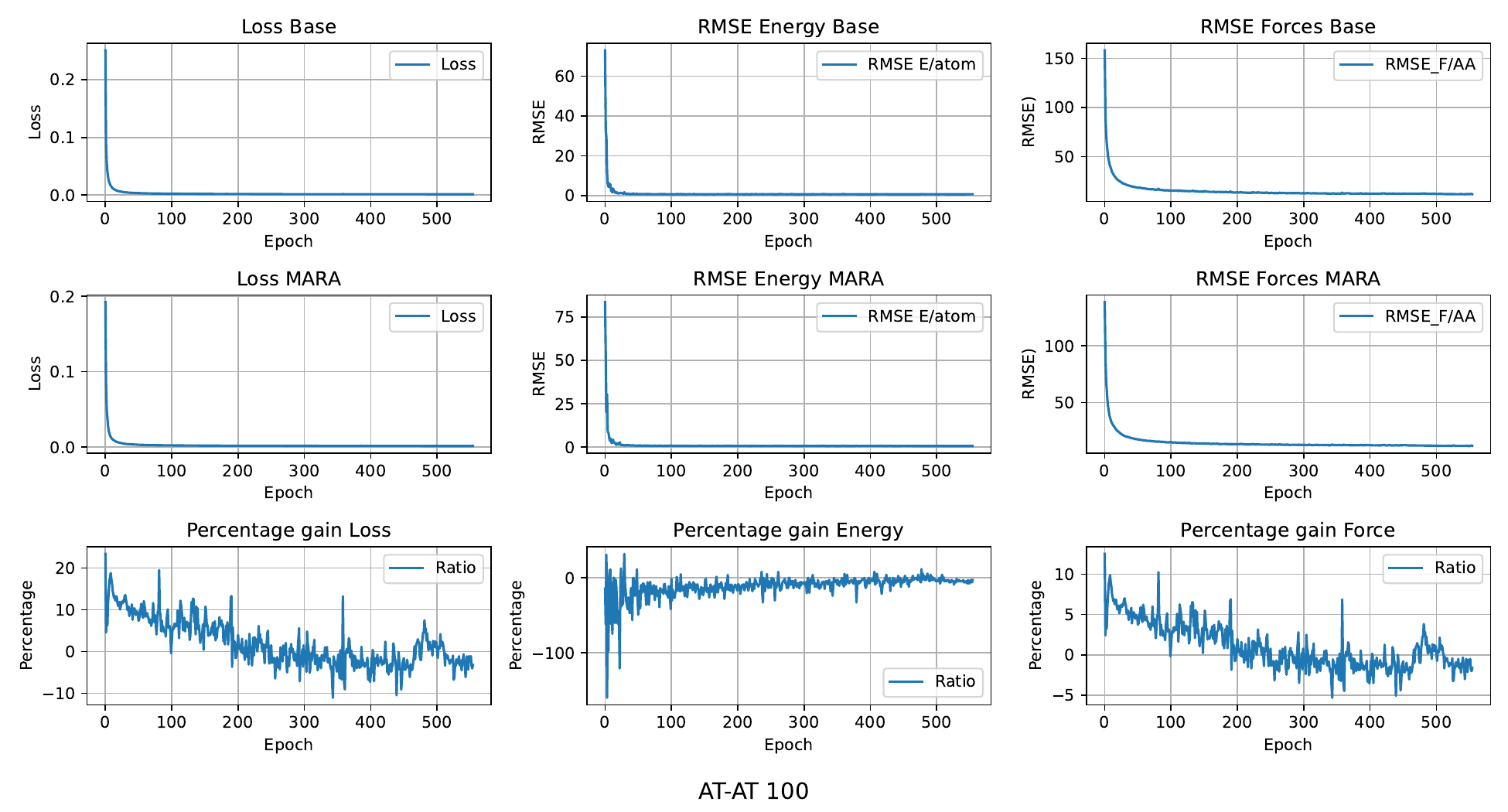}}
		\caption{Performance of models on the validation set on AT-AT with 100 samples. The subplots show, from left to right, Loss, Energy, and Strength; from top to bottom, Baseline, \modulenameShort, and the percentage ratio.}
		\label{img:ATAT100}
	\end{center}
\end{figure}

\begin{figure}[H]
	\vskip 0.2in
	\begin{center}
		\centerline{\includegraphics[width=0.9\linewidth]{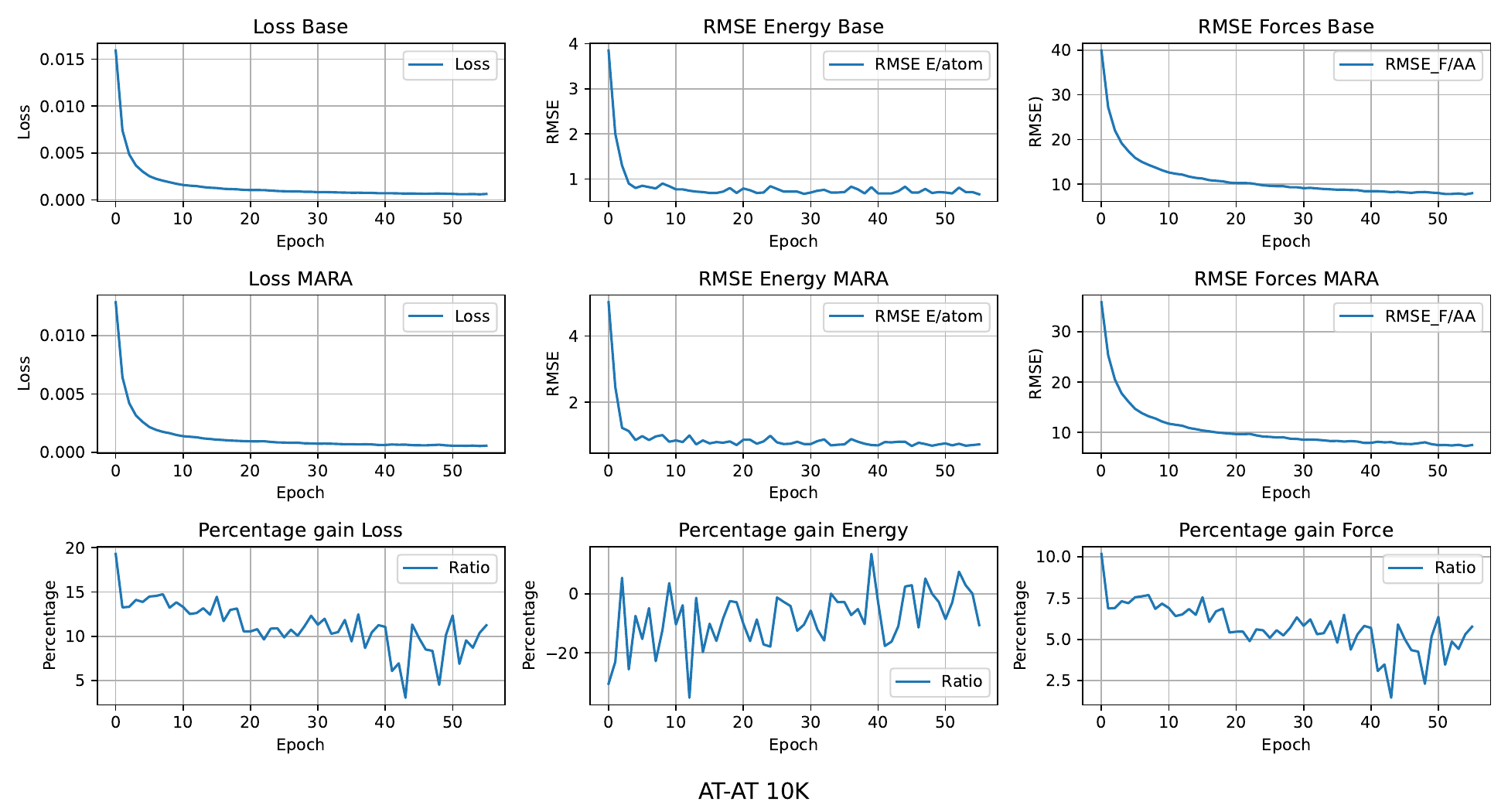}}
		\caption{Performance of models on the validation set on AT-AT with 10,000 samples. The subplots show, from left to right, Loss, Energy, and Strength; from top to bottom, Baseline, \modulenameShort, and the percentage ratio.}
		\label{img:ATAT10000}
	\end{center}
\end{figure}

Figures \ref{img:ATAT100} and \ref{img:ATAT10000} show that, for AT–AT pairs, \maceModulename yields worse force predictions than the baseline when trained on 100 samples, while outperforming it in the 10k-sample regime. In both settings, the module initially underperforms the baseline; however, with 100 samples, despite a substantial improvement in energy prediction during training, the performance ratio never becomes positive. In contrast, with 10,000 samples, training enables \maceModulename to match the baseline performance on energies.

Regarding forces, an opposite trend is observed. \maceModulename starts with a positive ratio, which decreases sharply in the 100-sample regime, whereas it remains more stable when trained on 10,000 samples. This behavior results in degraded test-set performance in the low-data regime and a clear overall improvement when sufficient training data are available.

\section{Performance analysis during molecular dynamics}\label{label:MD_traj}

To verify that the introduced module remains useful for molecular dynamics (MD) prediction and does not compromise the stability of the base model, we perform a short MD simulation using the Atomic Simulation Environment (ASE) framework~\cite{ISI:000175131400009, ase-paper}. The test system is the capped peptide Ac-Ala$_3$-NHMe. Initial velocities are sampled from the Maxwell–Boltzmann distribution at 500 K, and the simulation runs in the NVT ensemble using a Langevin thermostat with a friction coefficient of 0.1. The integration time step is 1 fs, and the trajectory is propagated for 10 000 fs (10 ps).

\begin{figure*}[t]
    \centering
    \begin{subfigure}{\textwidth}
        \centering
        \includegraphics[width=0.9\linewidth]{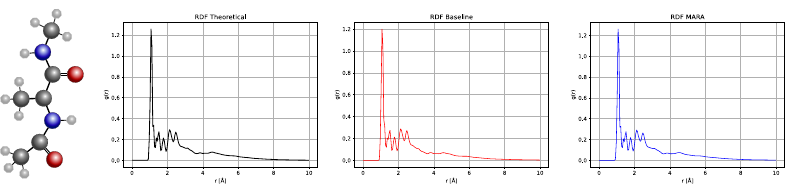}
        \caption{Radial distribution function (RDF) computed from the MD22 reference dataset, on MACE and MARA computed on MD simulations}
        \label{fig:app_MD_RDF}
    \end{subfigure}
    \hfill
    \vskip 0.1in
    \begin{subfigure}{\textwidth}
        \centering
        \includegraphics[width=\linewidth]{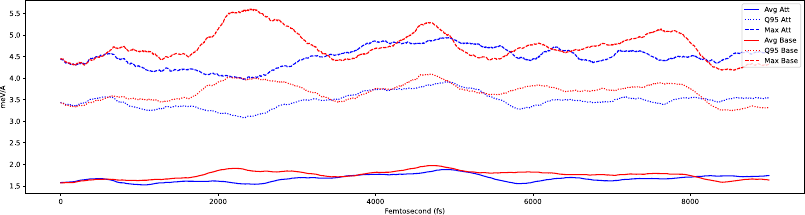}
        \caption{Molecular dynamics on Ac-Ala$_3$-NHMe}
        \label{fig:app_MD_run}
    \end{subfigure}
    \caption{MD validation. (a) RDF comparison of MACE and MACE-MARA with the MD22 reference. (b) Force magnitude statistics (mean, 95th percentile, maximum) during a 10 ps MD run on Ac-Ala$_3$-NHMe (1 fs timestep, NVT Langevin at 500 K, moving average window 500 fs).}
    \label{fig:main}
\end{figure*}

In Figure~\ref{fig:app_MD_RDF}, we compare the radial distribution function (RDF) computed from the MD22 reference dataset with the RDF obtained from the 10 000 configurations generated during the simulation, for both MACE and MACE-MARA.

In Figure~\ref{fig:app_MD_run}, we report the evolution of the force magnitude acting on the atoms: at each time step we compute the mean, the 95th percentile, and the maximum norm of the force. Since per-step measurements are highly noisy, we apply a 500 fs moving average window to highlight the overall trend.

Both models reproduce an RDF that is consistent with the reference RDF (MD22), indicating that the introduced module does not significantly distort the structural distribution. From Figure~\ref{fig:app_MD_run}, the force trajectories of the two models largely overlap and no divergence or numerical instability is observed. Moreover, fluctuations in MACE-MARA are slightly reduced compared to the baseline MACE model, suggesting that the attention mechanism may act as an adaptive filter, in agreement with observations reported in previous works~\cite{DBLP:conf/iclr/VelickovicCCRLB18, DBLP:conf/iclr/Brody0Y22}.

Finally, the computational overhead introduced by the module is small: on an NVIDIA H100 GPU, the simulation with MACE ($\ell=2$, grid resolution $4\times 8$) takes 940 s, while MACE-MARA requires 1007 s, corresponding to an overhead of approximately 7.1\% (about half of the cost reported in Table~\ref{table:ablation_grid}). This additional validation suggests that the proposed spherical attention mechanism, when integrated into a state-of-the-art SE(3)-equivariant model such as MACE, still allows for realistic MD simulations.

\section{Plug and play in other models}\label{app:othersplug}

As noted in the main text, our model is plug-and-play and can be integrated into SE(3)-equivariant architectures without significant structural modifications, enabling rapid adoption. In the main text, we employ MACE \cite{Batatia2022mace}, a body-ordered tensor network widely used in molecular dynamics. However, this choice does not restrict the applicability of MARA to MACE-based models.

In this appendix, we demonstrate the integration of MARA into other SE(3)-equivariant frameworks. Specifically, we implement MARA within NequIP \cite{DBLP:journals/corr/abs-2101-03164}, which is a message-passing architecture operating on irreducible representations. We use the original NequIP framework \href{https://github.com/mir-group/nequip}{(https://github.com/mir-group/nequip)} and report results on the aspirin molecule from the MD17 dataset \cite{chmiela2017machine}. Finally, we apply MARA to the attention-based equivariant model VisNET \cite{wang2024enhancing}, evaluating performance on the aspirin molecule from the MD17 dataset, further demonstrating the generality of the proposed approach across diverse equivariant architectures.

\subsection{NequIP}

As done in MACE, MARA is inserted into the InteractionBlock. For the test, we set the cutoff radius to $5.0~\text{\AA}$. The NequIPGNNModel is configured with a Bessel encoding of 8 basis functions, which are not trainable, and a polynomial cutoff exponent of 6. The convolutional network comprises 3 interaction layers, with $\ell_{\max}=1$, parity enabled, and 32 features per node. The radial network consists of 2 layers with 64 hidden units each. Dataset statistics are used to normalize the model, including the average number of neighbors, per-type energy scales based on force RMS, and per-type energy shifts based on per-atom energy mean.

\begin{table}[H]
	\caption{Analyzing the performance of NequIP and NequIP-MARA on Aspirin from MD17 dataset.}
	\label{table:NequIPresults}
	\begin{center}
		\begin{small}
			\begin{sc}
				\begin{tabular}{lccr}
					\toprule
					Metric & NequIP & NequIP-MARA & $\Delta (\%)$ \\
                    \midrule
                    Forces MAE        & 0.0558 & 0.0499 & \signcolor{-10.57} \\
                    Forces MAX Error  & 1.0779 & 0.8810 & \signcolor{-18.27} \\
                    Forces RMSE       & 0.0760 & 0.0682 & \signcolor{-10.26} \\
                    \midrule
                    Atom MAE        & 0.0018 & 0.0012 & \signcolor{-33.33} \\
                    Atom MAX Error  & 0.0155 & 0.0121 & \signcolor{-21.94} \\
                    Atom RMSE       & 0.0022 & 0.0016 & \signcolor{-27.27} \\
                    \midrule
                    Total Energy MAE        & 0.0391 & 0.0269 & \signcolor{-31.20} \\
                    Total Energy MAX Error  & 0.3268 & 0.2542 & \signcolor{-22.22} \\
                    Total Energy RMSE       & 0.0478 & 0.0342 & \signcolor{-28.45} \\
                    \midrule
                    Weighted Sum    & 0.0475 & 0.0384 & \signcolor{-19.16} \\		

                    \bottomrule
					
				\end{tabular}
			\end{sc}
		\end{small}
	\end{center}
	\vskip -0.2in
\end{table}

Analyzing the performance, we observe that MARA consistently improves NequIP on Aspirin from the MD17 dataset. Table~\ref{table:NequIPresults} reports the results obtained on the test set at the end of training. In this case, the gains are considerably larger than those observed for MACE, which may result from suboptimal NequIP settings, whereas in the main text we used a highly expressive MACE model. Nevertheless, it is evident that MARA provides a substantial improvement even on a model different from MACE.

\begin{algorithm}[ht]
    \caption{Forward pass of NequIP}
    \label{alg:SE3LayerNequIP}
    \begin{algorithmic}[1]
        
        \STATE $x \gets \text{linear\_1}(x)$
        \STATE normalize $x$ via avg\_num\_neighbors
        \IF{not first layer}
            \STATE $x \gets \text{ghost\_exchange}(x)$
        \ENDIF

        \STATE $x, att \gets \text{MARA}(x, positions, edge\_index)$
        \STATE $x \gets \text{tp\_scatter}(x, edge\_attrs, edge\_weights, edge\_src, edge\_dst)$
        \STATE $x \gets \text{linear\_2}(x)$
        
        \STATE $data[\text{NODE\_FEATURES\_KEY}] \gets x$
        
    \end{algorithmic}
\end{algorithm}

\subsection{VisNET}

To insert MARA into VisNET \href{https://github.com/ElwynWang/ViSNet/tree/main}{(https://github.com/ElwynWang/ViSNet/tree/main)}, it is necessary to adjust ViSNetBlock and the related Message Passing (ViS MP). Node features are embedded into a latent space of dimension 256 and processed through 5 stacked VisNET blocks, each incorporating spherical harmonics up to order $\ell_{max}$=2. Radial interactions are modeled using 32 exponential-normalized radial basis functions with a cutoff radius of 5.0 $\AA$. The attention mechanism uses 8 heads with SiLU activations and aggregates messages via summation. VisNET is trained using a combined energy–force loss, with forces dominating the objective, reflecting standard practice in molecular dynamics learning.

\begin{table}[H]
	\caption{Analyzing the performance of VisNET and VisNET-MARA on Aspirin from MD17 dataset.}
	\label{table:VisNETRes}
	\begin{center}
		\begin{small}
			\begin{sc}
				\begin{tabular}{lccr}
					\toprule
					Metric & VisNET & VisNET-MARA & $\Delta (\%)$ \\
                    \midrule
                    Scalar MAE  & 0.1850 & 0.1369 & \signcolor{-26.00} \\
                    Forces MAE  & 0.1679 & 0.1645 & \signcolor{-2.03} \\

                    \bottomrule
					
				\end{tabular}
			\end{sc}
		\end{small}
	\end{center}
	\vskip -0.1in
\end{table}

Table~\ref{table:VisNETRes} shows that the results on the aspirin test set from the MD17 data set demonstrate that MARA produces a significant performance improvement also in attention-based equivariant models.

We introduce the model as illustrated in pseudocode~\ref{alg:VisNETBlock}, treating MARA as a separate modular unit, and report the corresponding results in Table~\ref{table:VisNETRes}. We note that an alternative and equally natural integration of MARA within VisNET is shown in pseudocode ~\ref{alg:VisNETMessage}.

\begin{algorithm}[ht]
    \caption{Forward pass of VisNET block with MARA-style attention}
    \label{alg:VisNETBlock}
    \begin{algorithmic}[1]

        \STATE $x \gets \text{LayerNorm}(x)$

        \STATE $x, att \gets \text{MARA}(x, \text{pos}_{ij}, \text{edge\_index})$

        \STATE $q \gets \text{Linear}_q(x)$
        \STATE $k \gets \text{Linear}_k(x)$
        \STATE $v \gets \text{Linear}_v(x)$

        \STATE $d_k \gets \sigma(\text{Linear}_{dk}(f_{ij}))$
        \STATE $d_v \gets \sigma(\text{Linear}_{dv}(f_{ij}))$

    \end{algorithmic}
\end{algorithm}

\begin{algorithm}[ht]
    \caption{Message function of VisNET block with MARA-style attention}
    \label{alg:VisNETMessage}
    \begin{algorithmic}[1]

        \STATE $a_{ij} \gets \sum (q_i \odot k_j \odot d_k)$
        \STATE $a_{ij} \gets \sigma_{\text{attn}}(a_{ij}) \cdot \text{cutoff}(r_{ij})$

        \STATE $v_j \gets v_j \odot d_v$
        \STATE $v_j \gets v_j \odot a_{ij}$

        \STATE $v_j, att_{MARA} \gets \text{MARA}(v_j, \text{pos}_{ij}, \text{edge\_index})$

        \STATE $(s_1, s_2) \gets \text{Split}(\sigma(\text{Linear}_s(v_j)))$

    \end{algorithmic}
\end{algorithm}

\end{document}